\documentclass[twoside]{article}
\usepackage{aistats2016}

\usepackage[tableposition=top]{caption}
\usepackage[floatrowsep=quad]{floatrow}      
\newfloatcommand{capbtabbox}{table}[\captop][\FBwidth]

\usepackage{multirow}
\usepackage{pbox}
\usepackage{xcolor}

\usepackage[utf8]{inputenc}
\usepackage{graphicx}
\usepackage{amsmath}
\usepackage{amssymb}
\usepackage{amsthm}
\usepackage{bbm}        
\usepackage{subcaption}
\usepackage{algpseudocode}
\usepackage{algorithm}

\usepackage{booktabs}  
\newcommand{\ra}[1]{\renewcommand{\arraystretch}{#1}}

\usepackage{blindtext}

\usepackage{titlesec}   
\titlespacing\section{0pt}{7pt plus 2pt minus 2pt}{4pt plus 2pt minus 1pt}
\titlespacing\subsection{0pt}{6pt plus 2pt minus 2pt}{2pt plus 2pt minus 1pt}
\titlespacing\subsubsection{0pt}{5pt plus 2pt minus 2pt}{2pt plus 2pt minus 1pt}
\titlespacing\paragraph{0pt}{5pt plus 2pt minus 2pt}{2pt plus 2pt minus 1pt}

\usepackage{enumitem}
\setlist[itemize]{itemsep=0pt, topsep=0pt}
\setlist[enumerate]{itemsep=0pt, topsep=0pt}
\setlist[enumerate]{noitemsep, topsep=0pt}

\DeclareMathOperator{\trace}{tr}
\DeclareMathOperator{\prox}{prox}
\DeclareMathOperator{\diag}{diag}

\DeclareMathOperator{\st}{s.t.}
\newcommand{\tol}{\epsilon}
\DeclareMathOperator*{\minimize}{minimize}
\DeclareMathOperator*{\argmin}{argmin}

\DeclareMathOperator{\vectorform}{vectorform}
\DeclareMathOperator{\constant}{const.}
\newtheorem{proposition}{Proposition}

\newcommand{\ones}{\mathbf{1}}
\newcommand{\ind}[1]{\mathbbm{1}\{{#1}\}}
\newcommand{\tr}[1]{\trace\left({#1}\right)}
\newcommand{\vassilis}[1]{{\textcolor[rgb]{1,0,0}{#1}}}

\newcommand{\ER}{Erd\H{o}s R\'enyi}
\newcommand{\BA}{Barab\'asi-Albert}

\usepackage[style=authoryear-comp, backend=bibtex]{biblatex}

\begin{document}



\twocolumn[

\aistatstitle{How to learn a graph from smooth signals}

\aistatsauthor{ Vassilis Kalofolias }

\aistatsaddress{ Signal Processing Laboratory 2 (LTS2)\\ Swiss Federal Institute of Technology Lausanne (EPFL), Switzerland} ]


\begin{abstract}
We propose a framework that learns the graph structure underlying a set of smooth signals. Given $X\in\mathbb{R}^{m\times n}$ whose rows reside on the vertices of an unknown graph, we learn the edge weights $w\in\mathbb{R}_+^{m(m-1)/2}$ under the smoothness assumption that $\tr{X^\top LX}$ is small. 
We show that the problem is a weighted $\ell$-1 minimization that leads to naturally sparse solutions. We point out how known graph learning or construction techniques fall within our framework and propose a new model that performs better than the state of the art in many settings. We present efficient, scalable primal-dual based algorithms for both our model and the previous state of the art, and evaluate their performance on artificial and real data.
\end{abstract}

\section{INTRODUCTION}

We consider a matrix $X\in\mathbb{R}^{m\times n} = \left[x_1, \dots, x_m\right]^\top$, where each row $x_i\in\mathbb{R}^n$ resides on one of $m$ nodes of a graph $G$. Then each of the $n$ columns of $X$ can be seen as a signal on the same graph.
A simple assumption about data residing on graphs, but also the most widely used one is that it changes smoothly between connected nodes.
An easy way to quantify how smooth is a set of vectors $x_1, \dots, x_m\in\mathbb{R}^n$ on a given weighted undirected graph is through the function
\begin{equation*}
\frac{1}{2}\sum_{i,j}W_{ij}\|x_i-x_j\|^2,
\end{equation*}
where $W_{ij}$ denotes the weight of the edge between nodes $i$ and $j$. In words, if two vectors $x_i$ and $x_j$ from a smooth set reside on two well connected nodes (so $W_{ij}$ is big), they are expected to have a small distance. Using the graph Laplacian matrix $L = D-W$, where $D$ is the diagonal degree matrix with $D_{ii} = \sum_jW{ij}$, this function can be written in matrix form as
\begin{equation*}
\tr{X^\top LX}.
\end{equation*}
The importance of the graph Laplacian has long been known as a tool for embedding, manifold learning, clustering and semisupervised learning, see e.g. \textcite{belkin2001laplacian, zhu2003semi,belkin2006manifold}.
More recently we find an abundance of methods that exploit this notion of smoothness to regularize various machine learning tasks, solving problems of the form
\begin{equation}\label{eq:prob1}
\minimize_{X}~~g(X) + \tr{X^\top L X}.
\end{equation}
\textcite{zhang2006linear} use it to enhance web page categorization with graph information,  \textcite{zheng2011graph} for graph regularized sparse coding. \textcite{cai2011graph} use the same term to regularize NMF, \textcite{jiang2013graph} for PCA and \textcite{kalofolias2014matrix} for matrix completion.
Having good quality graphs is key to the success of the above methods.

\textit{The goal of this paper is to solve the complementary problem of learning a good graph:}
\begin{equation}\label{eq:prob2}
\minimize_{L\in{\cal L}}~~\tr{X^\top L X} + f(L),
\end{equation}
where $\cal L$ denotes the set of valid graph Laplacians.

\textit{Why is this problem important?} Firstly because it enables us to directly learn the hidden graph structure behind our data. Secondly because in most problems that can be written in the form of eq. \eqref{eq:prob1}, we are often given a noisy graph, or no graph at all. Therefore, starting from the initial graph and alternating between solving problems \eqref{eq:prob1} and \eqref{eq:prob2} we can at the same time get a better quality graph and solve the task of the initial problem.

\paragraph{Related Work.}
\textcite{dempster1972covariance} was one of the first to propose the problem of finding connectivity from measurements, under the name ``covariance selection''.
Years later, \textcite{banerjee2008model} proposed solving an $\ell$-1 penalized log-likelihood problem to estimate a sparse inverse covariance with unknown pattern of zeros.
However, while a lot of work has been done on inverse covariance estimation, the latter differs substantially from a graph Laplacian. For instance, the off-diagonal elements of a Laplacian must be non-positive, while it is not invertible like the  inverse covariance.

\textcite{wang2008label} learn a graph with normalized degrees by minimizing the objective
$\sum_i\|x_i-\sum_jw_{ij}x_j\|^2$, but they assume a fixed k-NN edge pattern. \textcite{daitch2009fitting} considered the similar objective $\|LX\|_F^2$ and they approximately minimized it with a greedy algorithm and a relaxation.


\textcite{zhang2010transductive} alternate between problems \eqref{eq:prob1} and a variation of \eqref{eq:prob2}. However, while they start from an initial graph Laplacian $L$, they finally learn a s.p.s.d. matrix that is not necessarily a valid Laplacian.
 
The works most relevant to ours are the ones by \textcite{lake2010discovering} and by \textcite{dong2015learning,dong2015laplacian}. In the first one, the authors consider a problem similar to the one of the inverse covariance estimation, but impose additional constraints in order to obtain a valid Laplacian. However, their final objective function contains many constraints and a computationally demanding log-determinant term that makes it difficult to solve. To the best of our knowledge, there is no scalable algorithm in the literature to solve their model. \textcite{dong2015learning} propose a model that outperforms the one by \citeauthor{lake2010discovering}, but still do not provide a scalable algorithm. This work is complementary to theirs, as we not only compare against their model, but also provide an analysis and a scalable algorithm to solve it.

\paragraph{Contributions.}
In this paper we make the link between smoothness and sparsity. We show that the smoothness term can be equivalently seen as a weighted $\ell$-1 norm of the adjacency matrix, and minimizing it leads to naturally sparse graphs (Section \ref{sec:laplacian}). Based on this, we formulate our objective as a weighted $\ell$-1 problem that we propose as a general framework for solving problem \eqref{eq:prob2}. Using this framework we propose a new model for learning a graph. We prove that our model has effectively one parameter that controls how sparse is the learnt graph (Section \ref{sec:learning}).

We show how our framework includes the standard Gaussian kernel weight construction, but also the model by \textcite{dong2015learning}. We simplify their model and prove fundamental properties (Section \ref{sec:learning}).

We provide a fast, scalable and convergent primal-dual algorithm to solve our proposed model, but also the one by \citeauthor{dong2015learning}. To the best of our knowledge, these are the first scalable solutions in the literature to learn a graph under smoothness assumption \eqref{eq:prob2} (Section \ref{sec:optimization}).

To evaluate our model, we first review different definitions of smooth signals in the literature. We show how they can be unified under the notion of graph filtering (Section \ref{sec:smooth_signals}).
We compare the models under artificial and real data settings. We conclude that our model is superior in many cases and achieves better connectivity when sparse graphs are sought (Section \ref{sec:experiments}).


\section{PROPERTIES OF THE LAPLACIAN}\label{sec:laplacian}

Throughout this paper we use the combinatorial graph Laplacian defined as $L=D-W$, where $D=\diag(W\ones)$ and $\ones = [1,\dots, 1]^\top$. The space of all valid combinatorial graph Laplacians, is by definition
\begin{align*}
{\cal L} = &\Big\{L\in\mathbb{R}^{m\times m}:\\
&\left(\forall i\neq j\right)~~~ L_{ij} = L_{ji}\leq0, ~~L_{ii} = -\sum_{j\neq i} L_{ij}\Big\}.
\end{align*}
In order to learn a valid graph Laplacian, we might be tempted to search in the above space, as is done e.g. by \textcite{lake2010discovering, dong2015learning}. We argue that it is more intuitive to search for a valid weighted adjacency matrix $W$ from the space
\begin{align*}
{\cal W}_m = \left\{W\in\mathbb{R}^{m\times m}_+: ~~ W = W^\top,~~ \diag(W)=0\right\},
\end{align*}
leading to simplified problems.
Even more, when it comes to actually solving the problem by optimization techniques, we should consider the space of all valid edge weights for a graph
\begin{align*}
{\cal W}_v = \left\{w\in\mathbb{R}^{m(m-1)/2}_+\right\},
\end{align*}
so that we do not have to deal with the symmetricity of $W$ explicitly. The spaces ${\cal L}$, ${\cal W}_m$ and ${\cal W}_v$ are equivalent, and connected by bijective linear mappings. In this paper we use ${\cal W}_m$ to analyze the problem in hand and ${\cal W}_v$ when we solve the problem. Table \ref{tab:vectorform} exhibits some of the equivalent forms in the three spaces.

\begin{table}
\caption{Equivalent terms for representations from sets ${\cal L}, {\cal W}_m, {\cal W}_v$. We use $z = \vectorform(Z)$, and linear operator $S$ that performs summation in the vector form.}
\label{tab:vectorform}
\centering
\resizebox{\columnwidth}{!}{%
\footnotesize
\begin{tabular}{@{}lclcl@{}}
\toprule
$L\in{\cal L}$ && $W\in{\cal W}_m$ && $w\in{\cal W}_v$\\
\midrule
$2\tr{X^\top LX}$ && $\|W\circ Z\|_{1,1}$ && $\displaystyle 2w^\top z$ \\
$\tr{L}$ && $\|W\|_{1,1}$ && $2w^\top \ones = 2\|w\|_1$ \\
-- && $\|W\|_F^2$ && $2\|w\|^2_2$ \\
$\diag(L)$  &&  $W\ones$ && $Sw$ \\
$\ones^\top \log(\diag(L))$  &&  $\ones^\top \log(W\ones)$ && $\displaystyle \ones^\top \log(Sw)$ \\
$\|L\|_F^2$   &&   $\|W\|_F^2+\|W\ones\|^2_2$ && $2\|w\|^2_2 + \|Sw\|^2_2$ \\
\bottomrule
\end{tabular}
}
\end{table}

\subsection{Smooth manifold means graph sparsity}
Let us define the \textit{pairwise distances matrix} $Z\in\mathbb{R}_+^{m\times m}$:
\[Z_{i,j} = \left\|x_i-x_j\right\|^2.\]
Using this, we can rewrite the trace term as
\begin{align}\label{eq:smoothness_sparsity}
\tr{X^\top LX} = \frac{1}{2}\tr{WZ} = \frac{1}{2}\left\|W\circ Z\right\|_{1,1},
\end{align}
where $\|A\|_{1,1}$ is the elementwise norm-1 of $A$ and $\circ$ is the Hadamard product (see Appendix).
In words, \textit{the smoothness term is a weighted $\ell$-1 norm of $W$}, encoding \textit{weighted sparsity}, that penalizes edges connecting distant rows of $X$. The interpretation is that when the given distances come from  a smooth manifold, the corresponding graph has a sparse set of edges, preferring only the ones associated to small distances in $Z$.

Explicitly adding a sparsity term $\gamma\|W\|_{1,1}$ to the objective function is a common tactic for inverse covariance estimation. However, it brings little to our problem, as here it can be translated as merely adding a constant to the squared distances in $Z$:
\begin{equation}\label{eq:useless_sparsity}
\tr{X^\top LX} + \gamma\|W\|_{1,1} = \frac{1}{2}\left\|W\circ (2\gamma+Z)\right\|_{1,1}.
\end{equation}

Note that all information of $X$ conveyed by the trace term is contained in the pairwise distances matrix $Z$, so that the original could be omitted. Moreover, using the last term of eq. \eqref{eq:smoothness_sparsity} instead of the trace enables us to define other kinds of distances instead of Euclidean.

Note finally that the separate rows of $X$ do not have to be smooth signals in some sense. Two non-smooth signals $x_i, x_j$ can have a small distance between them, and therefore a small entry $Z_{i,j}$.


\section{WHAT IS A SMOOTH SIGNAL?} \label{sec:smooth_signals}
Given a graph, different definitions of what is a smooth signal have been used in different contexts. In this section we unify these different definitions using the notion of filtering on graphs. For more information about signal processing on graphs we refer to the work of \textcite{shuman2013emerging}. Filtering of a graph signal $x\in\mathbb{R}^m$ by a filter $h(\lambda)$ is defined as the operation\footnote{We denote by $h$ both the function $h:\mathbb{R}\rightarrow\mathbb{R}$ and its matrix counterpart $h:\mathbb{R}^{m\times m}\rightarrow\mathbb{R}^{m\times m}$ acting on the matrix's eigenvalues.} 
\begin{equation}\label{eq:filtering}
y = h(L)x = \sum_iu_ih(\lambda_i)u_i^\top x=\sum_iu_ih(\lambda_i)\hat x_i,
\end{equation}
where $\{u_i, \lambda_i\}$ are eigenvector-eigenvalue pairs of $L$, and $\hat x\in\mathbb{R}^m$ is the graph Fourier representation of $x$ containing its \textit{graph frequencies} $\hat x_i\in\mathbb{R}$. Low frequencies correspond to small eigenvalues, and low-pass or smooth filters correspond to decaying functions $h$. 

In the sequel we show how different models for smooth signals in the literature can be written as smoothing problems of an initial non-smooth signal.
We give an example of three different filters applied on the same signal in Figure \ref{fig:non_uniform_signals} (Appendix).

\vspace{-5pt}
\subsubsection*{Smooth signals by Tikhonov regularization.}
\vspace{-6pt}
Solving problem \eqref{eq:prob1} leads to smooth signals. By setting $g(x) = \frac{1}{\alpha}\|x-x_0\|^2$ we have a Tikhonov regularization problem, that given an arbitrary $x_0$ as input gives its graph-smooth version
$x=(\alpha L+I)^{-1}x_0$.
Equivalently, we can see this as filtering  $x_0$ by
\begin{equation}\label{eq:smooth_opt}h(\lambda) = \frac{1}{1+\alpha \lambda},\end{equation}
where big $\alpha$ values result in smoother signals. 

\vspace{-5pt}
\subsubsection*{Smooth signals from a probabilistic generative model.}
\vspace{-6pt}
\textcite{dong2015laplacian} proposed that smooth signals can be generated from a colored Gaussian distribution as $x = \bar x + \sum_iu_i \hat x_i$, 
where $\hat x_i \sim{\cal N}\left(0, \lambda_i^\dagger\right)$ and $\dagger$ denotes the pseudoinverse. 
Therefore $x$ follows the distribution
\begin{equation*}
x\sim{\cal N}\left(\bar x, L^\dagger\right).
\end{equation*}
To sample from the above, it suffices to draw an initial non-smooth signal $x_0\sim{\cal N}\left(0, I\right)$ and then compute
\begin{equation}\label{eq:gen_model}
x = \bar x + h(L)x_0,
\end{equation}
with $h(L) = \sqrt{L^\dagger}$, or equivalently filter it by \begin{equation}\label{eq:gen_model_filter}
h(\lambda)=\begin{cases}
\sqrt{{\lambda}^{-1}}& , \lambda>0\\
0&, \lambda = 0\end{cases}
\end{equation}
and add the mean $\bar x$. We point out here that using eq. \eqref{eq:gen_model} on any $x_0\sim{\cal N}\left(0, I\right)$ and for any filter $h(\lambda)$ would yield samples from 
\begin{equation*}
x\sim{\cal N}\left(\bar x, h(L)^2\right),
\end{equation*}
therefore the probabilistic generative model can be used for any filter $h$. However, it does not cover cases where the initial $x_0$ is not white Gaussian.

\vspace{-5pt}
\subsection*{Smooth signals by heat diffusion on graphs}\vspace{-6pt}
Another type of smooth signals in the literature results from the process of heat diffusion on graphs. See for example the work by \textcite{zhang2008graph} for an application on image denoising by heat diffusion smoothing on the pixels graph. Given an initial signal $x_0$, the result of the heat diffusion on a graph after time $t$ is 
$x = \exp(-Lt)x_0$, 
therefore the corresponding filter is
\begin{equation}\label{eq:opt_heat}
h(\lambda) = \exp(-t\lambda),
\end{equation}
where bigger values of $t$ result in smoother signals.

\section{LEARNING A GRAPH FROM SMOOTH SIGNALS}\label{sec:learning}
In order to learn a graph from smooth signals, we propose, as explained in Section \ref{sec:laplacian}, to rewrite problem \eqref{eq:prob2} using the weighted adjacency matrix $W$ and the pairwise distance matrix $Z$ instead of $X$:
\begin{equation}\label{eq:prob3}
\minimize_{W\in {\cal W}_m}~~~\|W\circ Z\|_{1,1} ~+~ f(W).
\end{equation}
Since $W$ is positive we could replace the first term by $\tr{WZ}$, but we prefer this notation to keep in mind that our problem already has a sparsity term on $W$. \textit{This means that $f(W)$ has to play two important roles: (1) prevent $W$ from going to the trivial solution $W=0$ and (2) impose further structure using prior information on $W$}. This said, depending on $f$ the solution is expected to be sparse, that is important for large scale applications.

In order to motivate this general graph learning framework, we show that the most standard weight construction, as well as the state of the art graph learning model are special cases thereof.

\subsection{Classic Laplacian computations}
In the literature one of the most common practices is to construct edge weights given $X$ from the Gaussian function 
\begin{equation}\label{eq:exp_d2}
w_{ij} = \exp\left(-\frac{\|x_i-x_j\|_2^2}{2\sigma^2}\right).
\end{equation}

It turns out that this choice of weights can be seen as the result of solving problem \eqref{eq:prob3} with a specific prior on the weights $W$:
\begin{proposition}The solution of the problem
\begin{align*}
\minimize_{W\in{\cal W}_m}  ~~ \|W\circ Z\|_{1,1} ~+~ 2\sigma^2\sum_{ij} W_{ij}\left(\log(W_{ij})-1\right)
\end{align*}
is given by eq. \eqref{eq:exp_d2}.
\end{proposition}
\vspace{-10pt}
\begin{proof}
The problem is edge separable and the objective can be written as 
$\sum_{i,j}W_{ij}Z_{ij} + 2\sigma^2W_{ij}(\log(W_{ij}-1))$. Deriving w.r.t. $W_{ij}$ we obtain the optimality condition $Z_{ij} + 2\sigma^2\log(W_{ij})=0$, or $W_{ij} = \exp(-Z_{ij}/(2\sigma^2))$, that proves the proposition.
\end{proof}
\vspace{-7pt}

Note that here, the logarithm in $f$ prevents the weights from going to $0$, leading to full matrices, and sparsification has to be imposed explicitly afterwards.

\subsection{Our proposed model}
Based on our framework \eqref{eq:prob3} our goal is to give a general purpose model for learning graphs, when no prior information is available. 
In order to obtain meaningful graphs, we want to make sure that \textit{each node has at least one edge with another node}. It is also desirable to \textit{have control of how sparse is the resulting graph}. To meet these expectations, we propose the following model with parameters $\alpha>0$ and $\beta\geq0$ controlling the shape of the edges:
\begin{equation}\label{eq:our_model}
\minimize_{W\in{\cal W}_m} ~\|W\circ Z\|_{1,1} ~-~ \alpha \ones^\top \log(W\ones) ~+~ \beta \|W\|_F^2.
\end{equation}

The logarithmic barrier acts on the node degree vector $W\ones$, unlike the model of Proposition 1 that has a similar barrier on the edges. This means that it forces the degrees to be positive, but does not prevent edges from becoming zero.
This improves the overall connectivity of the graph, without compromising sparsity. Note however, that adding solely a logarithmic term ($\beta = 0$) leads to very sparse graphs, and changing $\alpha$ only changes the scale of the solution and not the sparsity pattern (Proposition 2 for $\beta=0$). For this reason, we add the third term.

We showed in eq. \eqref{eq:useless_sparsity} that adding an $\ell$-1 norm to control sparsity is not very useful. On the other hand, adding a Frobenius norm we penalize the formation of big edges but do not penalize smaller ones. This leads to more dense edge patterns for bigger values of $\beta$. An interesting property of our model is that even if it has two terms shaping the weights, if we fix the scale we then need to search for only one parameter:
\begin{proposition}\label{proposition2}
Let $F(Z, \alpha, \beta)$ denote the solution of our model \eqref{eq:our_model} for input distances $Z$ and parameters $\alpha$, $\beta$. Then the following property holds for any $\gamma>0$:
\begin{equation}
F\left(Z, \alpha, \beta\right) = \gamma F\left(Z, \frac{\alpha}{\gamma}, \beta\gamma\right) = \alpha F\left(Z, 1, \alpha\beta\right).
\end{equation}
\end{proposition}
\vspace{-10pt}
\begin{proof}
{See appendix.}\end{proof}
\vspace{-10pt}

This means that for example if we want to obtain a $W$ with a fixed scale $\|W\|=s$ (for any norm), we can solve the problem with $\alpha=1$, search only for a parameter $\beta$ that gives the desired edge density and then normalize the graph by the norm we have chosen.

The main advantage of our model over the method by \textcite{dong2015laplacian}, is that it promotes connectivity by putting a log barrier directly on the node degrees. Even for $\beta=0$, we obtain the sparsest solution possible, that assigns at least one edge to each node. In this case, the distant nodes will have smaller degrees (because of the first term), but still be connected to their closest neighbour similarly to a 1-NN graph.

\subsection{Fitting the state of the art in our framework}

\textcite{dong2015laplacian} proposed the following model for learning a graph:
\begin{align*}
&\minimize_{L\in{\cal L}} ~\tr{X^\top LX} + \alpha \|L\|_F^2,\\
& ~~~~\st, \qquad \tr{L}=s.
\end{align*}
Parameter $s>0$ controls the scale (\citeauthor{dong2015laplacian} set it to $m$), and parameter $\alpha\geq0$ controls the density of the solution. This formulation has two weaknesses. First, using a Frobenius norm on the Laplacian has a reduced interpretability: the elements of $L$ are not only of different scales, but also linearly dependent. Secondly, optimizing it is difficult as it has 4 constraints on $L$: 3 in order to constrain $L$ in space $\cal L$, and one to keep the trace constant. We propose to solve their model using our framework: Using transformations of Table \ref{tab:vectorform}, we obtain the equivalent simplified model
\begin{align}
&\minimize_{W\in{\cal W}_m} ~\|W\circ Z\|_{1,1} + \alpha \|W\ones\|^2 + \alpha \|W\|_F^2,\nonumber\\
& ~~~~\st, \qquad \|W\|_{1,1}=s.\label{eq:xiaowens_model}
\end{align}
Using this parametrization, solving the problem becomes much simpler, as we show in Section \ref{sec:optimization}. Note that for $\alpha=0$ we have a linear program that assigns weight $s$ to the edge corresponding to the smallest pairwise distance in $Z$, and zero everywhere else. On the other hand, setting $\alpha$ to big values, we penalize big degrees (through the second term), and in the limit $\alpha\rightarrow\infty$ we obtain a dense graph with constant degrees across nodes. We can also prove some interesting properties of \eqref{eq:xiaowens_model}:

\begin{proposition}\label{proposition3}Let $H(Z, \alpha, s)$ denote the solution of model \eqref{eq:xiaowens_model} for input distances $Z$ and parameters $\alpha$ and $s$. Then for $\gamma>0$ the following properties hold:
\begin{align}
&H(Z + \gamma, \alpha, s) = H(Z, \alpha, s)\label{eq:xiaowen_prop1}\\
&H(Z, \alpha, s) = \gamma H\left(Z, \alpha\gamma, \frac{s}{\gamma}\right) = sH\left(Z, \alpha s, 1\right)\label{eq:xiaowen_prop2}
\end{align}
\end{proposition}
\vspace{-10pt}
\begin{proof}{See appendix.}
\end{proof}
\vspace{-7pt}

In other words, model \eqref{eq:xiaowens_model} is invariant to adding any constant to the squared distances. The second property means that similarly to our model, the scale of the solution does not change the shape of the connectivity. If we fix the scale to $s$, we obtain the whole range of edge shapes given by $H$ 
only by changing parameter $\alpha$.


\section{OPTIMIZATION}\label{sec:optimization}
An advantage of using the formulation of problem \eqref{eq:prob3} is that it can be solved efficiently for a wide range of choices of $f(W)$. We use primal dual techniques that scale, like the ones reviewed by \textcite{komodakis2014playing} to solve the two state of the art models: the one we propose and the one by \textcite{dong2015learning}. Using these as examples, it is easy to solve many interesting models from the general framework \eqref{eq:prob3}.  

In order to make optimization easier, we use the vector form representation from space ${\cal W}_v$ (see Table \ref{tab:vectorform}), so that the symmetricity does not have to be imposed as a constraint. We write the problem as a sum of three functions in order to fit it to primal dual algorithms reviewed by \textcite{komodakis2014playing}.
The general form of our objective is
\begin{align}
&\minimize_{w\in{\cal W}_v} ~f_1(w) + f_2(Kw) + f_3(w), \label{eq:opt_general}
\end{align}
where $f_1$ and $f_2$ are functions for which we can efficiently compute proximal operators, and $f_3$ is differentiable with gradient that has Lipschitz constant $\zeta \in (0, \infty)$. $K$ is a linear operator, so $f_2$ is defined on the dual variable $Kw$. In the sequel we explain how this general optimization framework can be applied to the two models of interest, leaving the details in the Appendix. For a better understanding of primal dual optimization or proximal splitting methods we refer the reader to the works of \textcite{combettes2011proximal, komodakis2014playing}.

In our model, the second term acts on the degrees of the nodes, that are a linear function of the edge weights. Therefore we use $K=S$, where $S$ is the linear operator that satisfies $W\ones = Sw$ if $w$ is the vectorform of $W$. In the first term we group the positivity constraint of ${\cal W}_v$ and the weighted $\ell$-1, and the second and third terms are the priors for the degrees and the edges respectively. In order to solve our model we define
\vspace{-10pt}
\begin{align*}
&f_1(w) = \ind{w\geq 0} + 2w^\top z,\\
&f_2(d) = -\alpha \ones^\top \log(d),\\
&f_3(w) = \beta \|w\|^2,\text{ with }\zeta = 2\beta,
\end{align*}
where $\ind{}$ is the indicator function that becomes zero when the condition in the brackets is satisfied, infinite otherwise. Note that the second function $f_2$ is defined on the dual variable $d = Sw\in\mathbb{R}^m$, that here is very conveniently the vector of the node degrees. 

\begin{algorithm}
\caption{Primal dual algorithm for model \eqref{eq:our_model}.}
\label{alg:ours}
\begin{algorithmic}[1]
\footnotesize
\State \textbf{Input:} $z,\alpha, \beta$, $w^0\in{\cal W}_v$, $d^0\in\mathbb{R}^m_+$, $\gamma$, tolerance $\tol$
\For{$i = 1,\dots,i_{max}$}
\State $ y^i = w^i - \gamma (2\beta w^i + S^\top d^i)$
\State $\bar y^i = d^i + \gamma (Sw^i)$
\State $p^i = \max(0, y^i-2\gamma z)$
\State $\bar p^i = (\bar y^i - \sqrt{(\bar y^i)^2+4\alpha\gamma})/2$
\Comment{elementwise}
\State $q^i = p^i - \gamma (2\beta p^i + S^\top p^i)$
\State $\bar q^i = \bar p^i + \gamma (Sp^i)$
\State $w^i = w^i - y^i + p^i$;
\State $d^i = d^i - \bar y^i + \bar q^i$;
\If{$\|w^i-w^{i-1}\|/\|w^{i-1}\|<\tol$ \textbf{and} \\
\qquad\quad $\|d^i-d^{i-1}\|/\|d^{i-1}\|<\tol$}
\State \textbf{break}
\EndIf
\EndFor
\end{algorithmic}
\end{algorithm}

For model \eqref{eq:xiaowens_model} we can define in a similar way
\begin{align*}
&f_1(w) = \ind{w\geq 0} + 2w^\top z,\\
&f_2(c) = \ind{c=s},\\
&f_3(w) = \alpha \left(2\|w\|^2+\|Sw\|^2\right),\text{ with }\zeta = 2\alpha(m+1),
\end{align*}
and use $K = 2\ones^\top$ so that the dual variable is $c = Kw = \|W\|_{1,1}$, constrained by $f_2$ to be equal to $s$.

Using these functions, the final algorithm for our model is given as Algorithm \ref{alg:ours}, and for the model by \citeauthor{dong2015laplacian} as Algorithm \ref{alg:xiaowens} in the Appendix. Vector $z\in\mathbb{R}_+^{m\times (m-1)/2}$ is the vector form of $Z$, and parameter $\gamma \in (0, ~1$$+$$\zeta$$+$$\|K\|)$ is the stepsize.

\subsection{Complexity and Convergence}
Both algorithms that we propose have a complexity of ${\cal O}(m^2)$ per iteration, for $m$ nodes graphs, and they can easily be parallelized. As the objective functions of both models are proper, convex, and lower-semicontinuous, our algorithms are guaranteed to converge to the minimum (\cite{komodakis2014playing}).


\section{EXPERIMENTS}\label{sec:experiments}
We compare our model against the state of the art model by \textcite{dong2015laplacian} solved by our Algorithm 2 for both artificial and real data. Comparing to the model by \textcite{lake2010discovering} was not possible even for the small graphs of our artificial experiments, as there is no scalable algorithm in the literature and the use of CVX with the log-determinant term is prohibitive. Other models based on the log-det term, for which scalable algorithms exist, are irrelevant to our problem as a sparse inverse covariance is not a valid Laplacian and are known to not perform well for our setting (see \textcite{dong2015laplacian} for a comparison).

\subsection{Artificial data}
The difficulty of solving problem \eqref{eq:prob3} depends both on the quality of the graph behind the data and on the type of smoothness of the signals. We test 4 different types of graphs using 3 different types of signals.

\begin{table*}[t]\centering
\footnotesize 
\caption{Performance of Different Models on Artificial Data.}
\label{tab:artificial}
\begin{tabular}{@{}rcccccccccccc@{}}\toprule
& & \multicolumn{3}{c}{Tikhonov} & \phantom{}& \multicolumn{3}{c}{Generative Model} & \phantom{} & \multicolumn{3}{c}{Heat Diffusion}\\ \cmidrule{3-5} \cmidrule{7-9} \cmidrule{11-13}
&& \textbf{base} & \textbf{\hspace{-2pt}Dong etal\hspace{-2pt}} & \textbf{Ours} &&  \textbf{base} & \textbf{\hspace{-2pt}Dong etal\hspace{-2pt}} & \textbf{Ours} &&  \textbf{base} & \textbf{\hspace{-2pt}Dong etal\hspace{-2pt}} & \textbf{Ours} \\
\midrule
\textbf{Rand. Geometric}\\
F-measure &&       0.685  & 0.885  & \textbf{0.913} && 0.686 & 0.877 & \textbf{0.909} && 0.758 & 0.837 & \textbf{0.849} \\
edge $\ell$-1 &&   0.866  & 0.357  & \textbf{0.298} && 0.798 & 0.371 & \textbf{0.348} && 0.609 & 0.524 & \textbf{0.447} \\
edge $\ell$-2 &&   0.676  & 0.376  & \textbf{0.336} && 0.658 & 0.397 & \textbf{0.390} && 0.576 & 0.531 & \textbf{0.468} \\
degree $\ell$-1 && 0.142  & 0.146  & \textbf{0.065} && 0.261 & 0.147 & \textbf{0.112} && 0.209 & 0.227 & \textbf{0.142} \\
degree $\ell$-2 && 0.708  & 0.172  & \textbf{0.079} && 0.689 & 0.174 & \textbf{0.128} && 0.474 & 0.264 & \textbf{0.176} \\
\textbf{Non Uniform}\\
F-measure &&       0.686  & \textbf{0.863}  & 0.858 && 0.633 & \textbf{0.840} & 0.832 && 0.766 & \textbf{0.839} & 0.830 \\
edge $\ell$-1 &&   0.821  & 0.423  & \textbf{0.349} && 0.864 & 0.487 & \textbf{0.472} && 0.594 & 0.565 & \textbf{0.473} \\
edge $\ell$-2 &&   0.706  & 0.434  & \textbf{0.344} && 0.735 & 0.480 & \textbf{0.474} && 0.550 & 0.587 & \textbf{0.451} \\
degree $\ell$-1 && 0.160  & 0.184  & \textbf{0.055} && 0.235 & 0.185 & \textbf{0.100} && 0.233 & 0.255 & \textbf{0.128} \\
degree $\ell$-2 && 0.612  & 0.209  & \textbf{0.073} && 0.632 & 0.215 & \textbf{0.161} && 0.427 & 0.324 & \textbf{0.157} \\
\textbf{\ER}\\
F-measure &&       0.288  & 0.766  & \textbf{0.893} && 0.199 & 0.755 & \textbf{0.896} && 0.377 & 0.629 & \textbf{0.655} \\
edge $\ell$-1 &&   1.465  & 0.448  & \textbf{0.391} && 1.566 & 0.478 & \textbf{0.427} && 1.379 & \textbf{0.832} & 0.841 \\
edge $\ell$-2 &&   1.060  & 0.442  & \textbf{0.402} && 1.105 & 0.457 & \textbf{0.440} && 1.033 & 0.735 & \textbf{0.726} \\
degree $\ell$-1 && 0.094  & 0.107  & \textbf{0.046} && 0.099 & 0.105 & \textbf{0.066} && 0.182 & \textbf{0.179} & 0.183 \\
degree $\ell$-2 && 0.986  & 0.161  & \textbf{0.066} && 1.312 & 0.181 & \textbf{0.151} && 0.892 & \textbf{0.236} & 0.273 \\
\textbf{\BA}\\
F-measure &&       0.345  & 0.710  & \textbf{0.868} && 0.382 & 0.739 & \textbf{0.838} && 0.352 & 0.690 & \textbf{0.765} \\
edge $\ell$-1 &&   1.531  & 0.614  & \textbf{0.533} && 1.496 & 0.652 & \textbf{0.624} && 1.468 & 0.740 & \textbf{0.675} \\
edge $\ell$-2 &&   1.061  & 0.568  & \textbf{0.506} && 1.036 & 0.611 & \textbf{0.571} && 1.041 & 0.662 & \textbf{0.590} \\
degree $\ell$-1 && 0.175  & 0.264  & \textbf{0.111} && \textbf{0.199} & 0.264 & 0.207 && 0.254 & 0.317 & \textbf{0.148} \\
degree $\ell$-2 && 0.554  & 0.340  & \textbf{0.201} && 0.556 & 0.333 & \textbf{0.287} && 0.568 & 0.414 & \textbf{0.283} \\
\bottomrule
\end{tabular}
\end{table*}

\vspace{-5pt}

\paragraph{Graph Types.}
We use two 2-D manifold based graphs, one uniformly and one non-uniformly sampled, and two graphs that are not manifold structured:
\begin{enumerate}
\item Random Geometric Graph (RGG): We sample $x$ uniformly from $[0,1]^2$ and connect nodes using eq. \eqref{eq:exp_d2} with $\sigma=0.2$, then threshold weights $<0.6$.
\item Non-uniform: We sample $x$ in $[0,1]\times[0,5]$  from a non-uniform distribution $p_{x_1,x_2}\propto 1/(1+\alpha x_2)$ and connect nodes using eq. \eqref{eq:exp_d2} with $\sigma=0.2$. We threshold weights smaller than the best connection of the most distant node ($\approx0.01$).
\item \ER: Random graph as proposed by \textcite{gilbert1959random} ($p$=$3/m$).
\item \BA: Random scale-free graph with preferential attachment as proposed by \textcite{barabasi1999emergence} ($m_0$=$1$, $m$=$2$).
\end{enumerate}
\vspace{-5pt}
\paragraph{Signal Types.}
 To create a smooth signal we filter a Gaussian i.i.d. $x_0$ by eq. \eqref{eq:filtering}, using one of the three filter types of Section \ref{sec:smooth_signals}. We normalize the Laplacian ($\|L\|_2=1$) so that the filters $g(\lambda)$ are defined for $\lambda\in[0,1]$. See Table \ref{tab:smooth_filters} (Appendix) for a summary.
\begin{enumerate}
\item Tikhonov: $g(\lambda) = \frac{1}{1+10\lambda}$ as in eq. \eqref{eq:smooth_opt}.
\item Generative Model: $g(\lambda) = 1/\sqrt{\lambda}$ if $\lambda>0$, $g(0)=0$ from model of eq. \eqref{eq:gen_model_filter} ($\bar x = 0$).
\item Heat Diffusion: $g(\lambda) = \exp(-10\lambda)$ as eq. \eqref{eq:opt_heat}.
\end{enumerate}
For all cases we use $m=100$ nodes, smooth signals of length $n = 1000$, and add $10\%$ ($\ell$-2 sense) noise before computing pairwise distances. We perform grid search to find the best parameters for each model. We repeat the experiment 20 times for each case and report the average result of the parameter value that performs best for each of the different metrics.
\vspace{-5pt}
\paragraph{Metrics.}
Since we have the ground truth graphs for each case, we can measure directly the relative edge error in the $\ell$-$1$ and $\ell$-$2$ sense. We also report the relative error of the weighted degrees $d_i = \sum_jW_{ij}$. This is important because both models are based on priors on the degrees as we show in section \ref{sec:learning}. We also report the F-measure (harmonic mean of edge precision and recall), that only takes into account the binary pattern of existing edges and not the weights.
\vspace{-5pt}
\paragraph{Baselines.}
The baseline for the relative errors is a classic graph construction using equation \eqref{eq:exp_d2} with a grid search for the best $\sigma$. Note that this exact equation was used to create the two first artificial datasets. However, using a fully connected graph with the F-measure does not make sense. For this metric the baseline is set to the best edge pattern found by thresholding \eqref{eq:exp_d2} with different thresholds. 

Table \ref{tab:artificial} summarizes all the results for different combinations of graphs/signals. In most of them, our model performs better for all metrics. We can see that the signals constructed following the generative model \eqref{eq:gen_model} do not yield better results in terms of graph reconstruction. Using smoother ``Tikhonov'' signals from eq. \eqref{eq:smooth_opt} or ``Heat Diffusion'' signals from \eqref{eq:opt_heat} by setting $\lambda=20$ yielded slightly worse results in both cases (not reported here). It also seems that the results are slightly better for the manifold related graphs than for the {\ER} and {\BA} models, an effect that is more prevalent when we use signals of length $n=100$ smooth signals instead of $1000$ (c.f. Table \ref{tab:artificial_100} of Appendix). This would be interesting to investigate theoretically.

\subsection{Real data}
We also evaluate the performance of our model on real data. In this case, the actual ground truth graph is not known. We therefore measure the performance of different models on spectral clustering and label propagation, two algorithms that depend solely on the graph. 
Note that an explicit Laplacian normalization is not needed for the learned models (it is even harmful as found experimentally), since this role is already played by the regularization.

\begin{figure*}
    \centering
    \begin{subfigure}[b]{0.32\textwidth}
        \includegraphics[scale=.32]{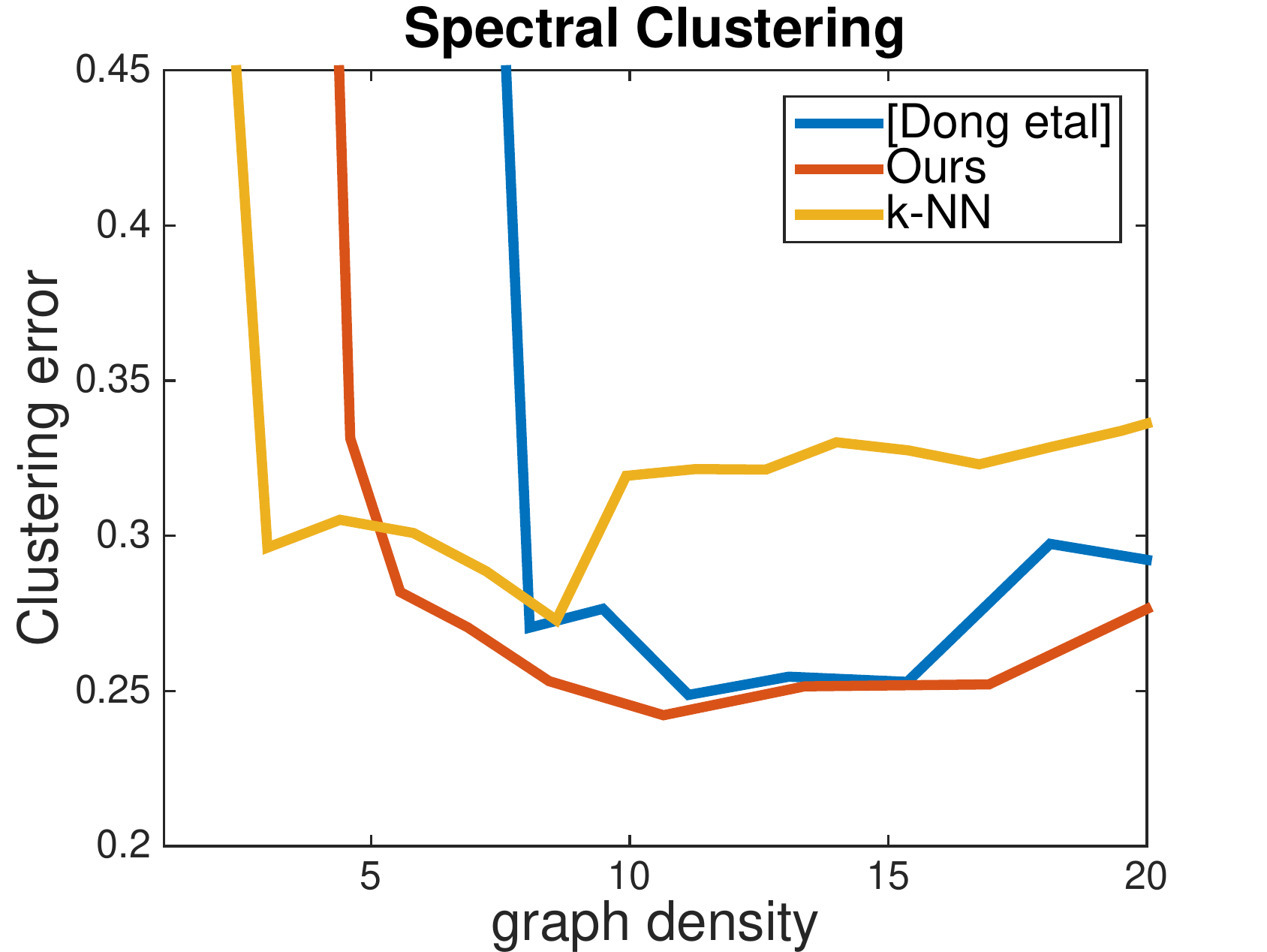}
    \end{subfigure}
    \begin{subfigure}[b]{0.32\textwidth}
        \includegraphics[scale=.32]{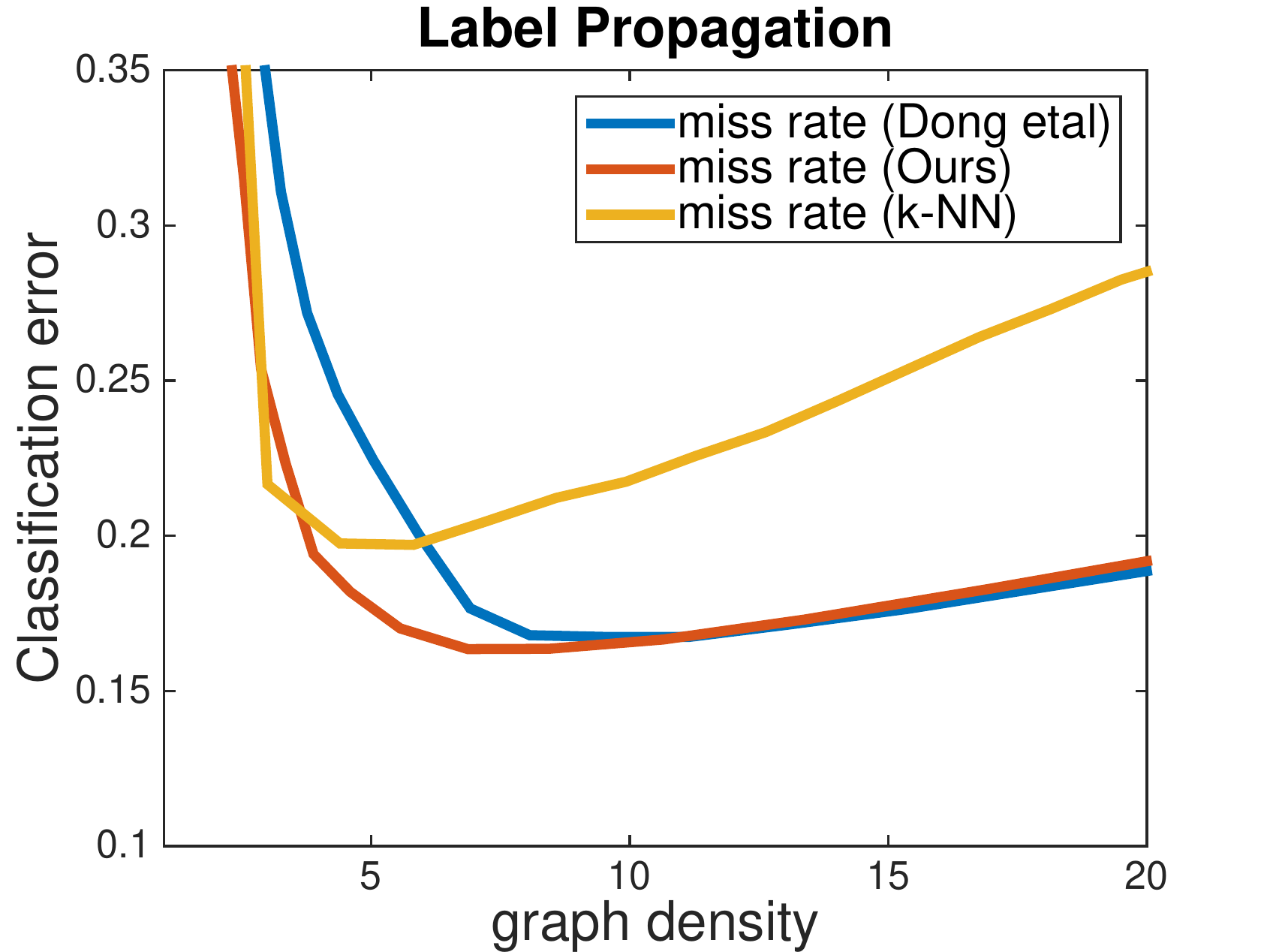}
    \end{subfigure}
    \begin{subfigure}[b]{0.32\textwidth}
        \includegraphics[scale=.32]{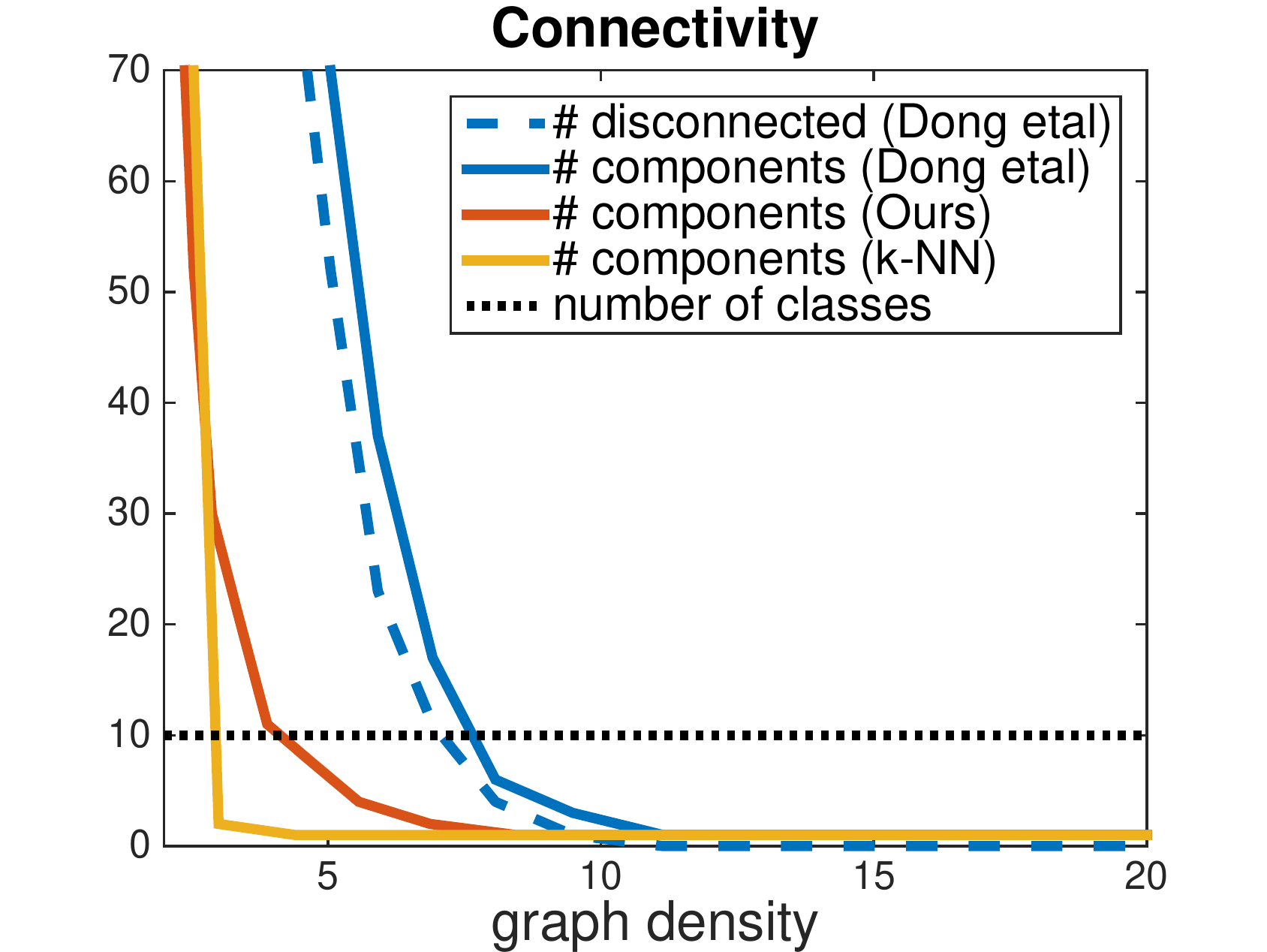}
    \end{subfigure}
    \caption{Graph learned from 1001 USPS images. \textbf{Left}: Clustering quality. \textbf{Middle}: Label propagation quality. \textbf{Right}: Number of completely disconnected components (continuous lines) and number of disconnected nodes for model by Dong etal. (blue dashed line). Our model and k-NN have no disconnected nodes.}\label{fig:USPS}
\end{figure*}
\begin{figure*}
    \centering
    \begin{subfigure}[b]{0.32\textwidth}
        \includegraphics[scale=.32]{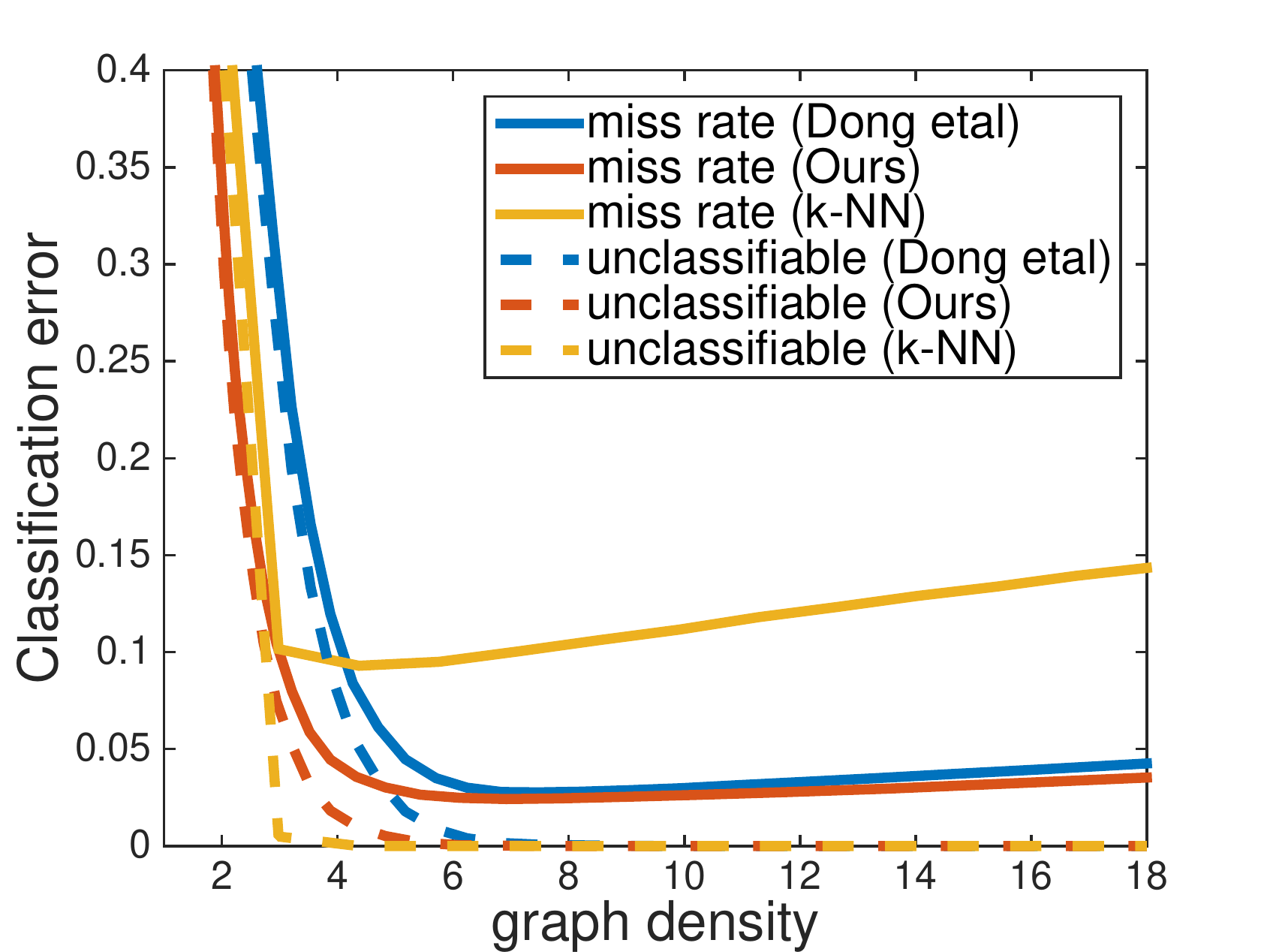}
    \end{subfigure}
    \begin{subfigure}[b]{0.32\textwidth}
        \includegraphics[scale=.32]{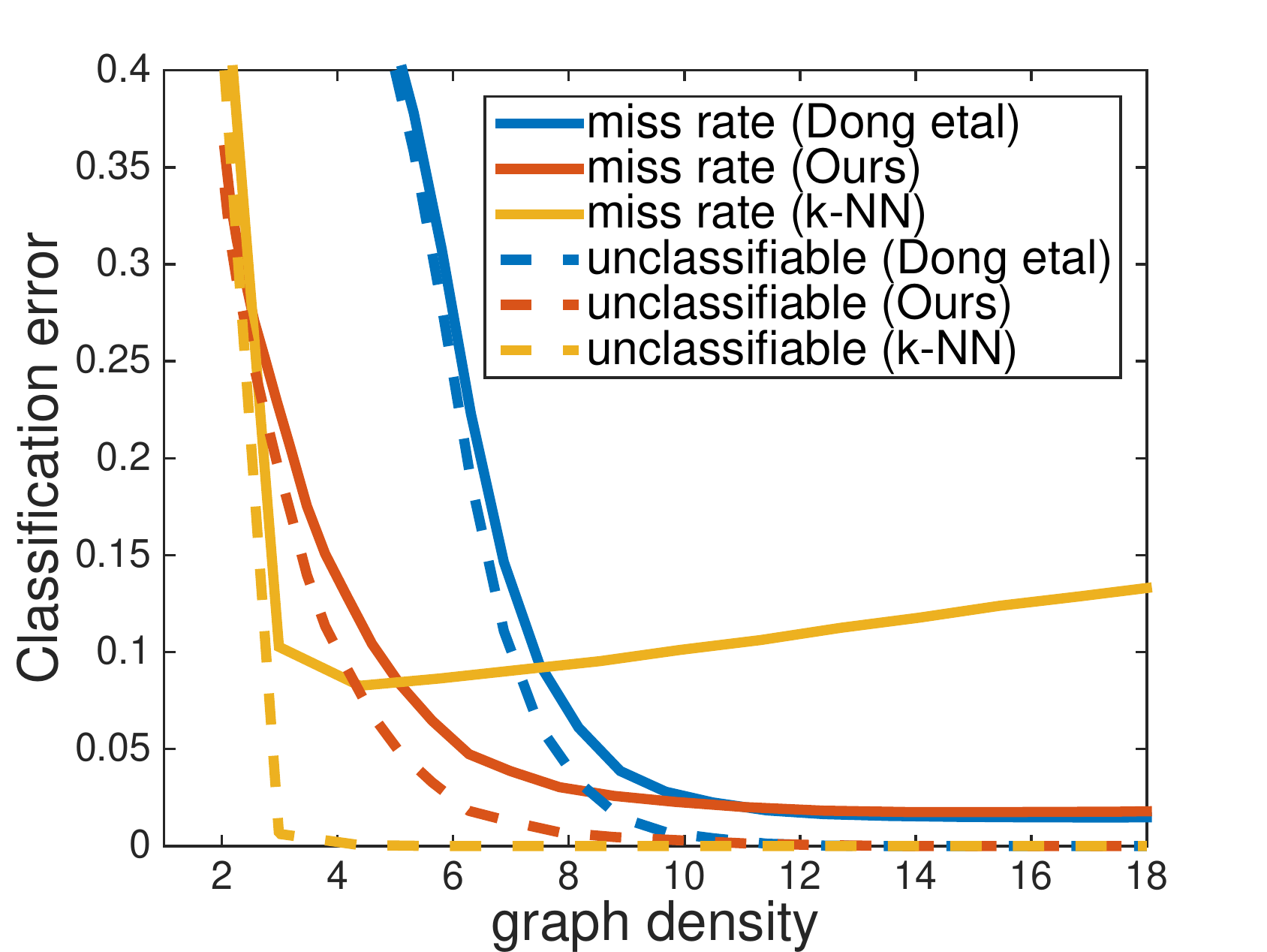}
    \end{subfigure}
    \begin{subfigure}[b]{0.32\textwidth}
        \includegraphics[scale=.32]{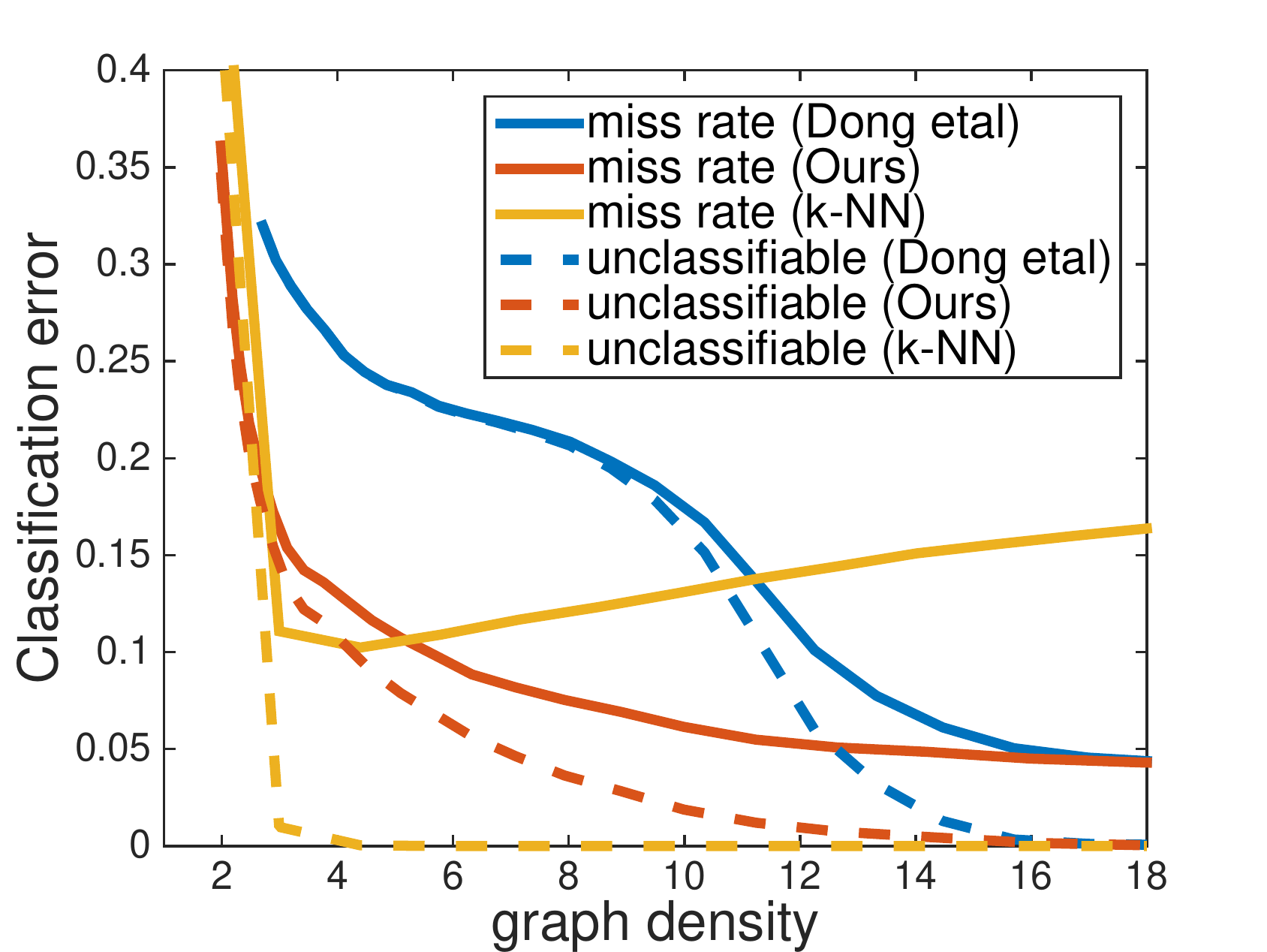}
    \end{subfigure}
    \caption{Label propagation for the problem ``1'' vs. ``2'' of MNIST with different class size proportions: 1 to 4 \textbf{(left)}, 1 to 1 \textbf{(middle)} or 4 to 1 \textbf{(right)}. Missclassification rate for different number of edges per node.}\label{fig:MNIST_lab_prop}
\end{figure*}

\vspace{-5pt}
\subsubsection*{Learning the graph of USPS digits}\vspace{-5pt}
We first learn the graph connecting 1001 different images of the USPS dataset, that are images of digits from 0 to 9 (10 classes). We follow \textcite{zhu2003semi} and sample the class sizes non-uniformly. For each class $i\in\{1\dots10\}$ we take round($2.6i^2$) images, resulting to classes with sizes from 3 to 260 images each. 
We learn graphs of different densities using both models. As baseline we use a k-Nearest Neighbors (k-NN) graph for different $k$.

For each of the graphs, we run standard spectral clustering (as in the work of \cite{ng2002spectral} but without normalizing the Laplacian) with k-means {100 times}. We also run label propagation we choose {100 times} a different subset of $10\%$ known labels.

In Fig. \ref{fig:USPS} we plot the behavior of different models for different density levels. The horizontal axis is the average number of non-zero edges per node. In the \textbf{left plot} we see the clustering quality. Even though the best result of both algorithms is almost the same (0.24 vs 0.25), our model is more robust in terms of the graph density choice.
A similar behavior is exhibited for label propagation plotted in the \textbf{middle}. The classification quality is better for our model in the sparser graph density levels. 

The robustness of our model for small graph densities can be explained by the connectivity quality plotted in the \textbf{right}. The continuous lines are the number of different connected components in the learned graphs, that is a measure of connectivity: the less components there are, the better connected is the graph. The dashed blue line is the number of disconnected nodes of model \cite{dong2015laplacian}. The latter fails to assign connections to the most distant nodes, unless the density of the graph reaches a fairly high level. If we want a graph with 6 edges per node, our model returns a graph with 3 components and no disconnected nodes. The model by \citeauthor{dong2015laplacian} returns a graph with 35 components out of which 22 are disconnected nodes.

Note that in real applications where the best density level is not known a priori, it is important for a graph learning model to perform well for sparse levels. This is especially the case for large scale applications, where more edges mean more computations.

\vspace{-5pt}
\paragraph{Time:}
Algorithm 1 implemented in Matlab\footnote{Code for both models is available as part of the open-source toolbox GSPBox by \textcite{perraudin2014gspbox} using code from UNLocBoX, \textcite{perraudin2014unlocbox}.} learned a 10-edge/node graph of 1001 USPS images in 5 seconds (218 iterations) and Algorithm 2 in 1 minute (2043 iterations) on a standard PC for tolerance $\epsilon = 1e$-4.

\vspace{-5pt}
\subsubsection*{Learning the graph of MNIST 1 vs 2}\vspace{-5pt}
To demonstrate the different behaviour of the two models for non-uniform sampling cases, we use the problem of classification between digits 1 and 2 of the MNIST dataset. This problem is particular because digits ``1'' are close to each other (average square distance of 45), while digits ``2'' differ more from each other (average square distance of 102). In Figure \ref{fig:MNIST_lab_prop} we report the average miss-classification rate for different class size proportions, with 40 1's and 160 2's \textbf{(left)}, 100 1's and 100 2's \textbf{(middle)} or  160 1's and 40 2's \textbf{(right)}. Results are averaged over 40 random draws. The dashed lines denote the number of nodes contained in components without labeled nodes, that can not be classified. In this case, the model of \cite{dong2015laplacian} fails to recover edges between different digits ``2'' unless the returned graph is fairly dense, unlike our model that even for very sparse graph levels treats the different classes more fairly. The effect is stronger when the set of 2's is also the smallest of the two.


\section{CONCLUSION}
We introduce a new way of addressing the problem of learning a graph under the assumption that $\tr{X^\top LX}$ is small. We show how the problem can be simplified into a weighted sparsity problem, that implies a general framework for learning a graph. We show how the standard Gaussian weight construction from distances is a special case of this framework. We propose a new model for learning a graph, and provide an analysis of the state of the art model of \textcite{dong2015laplacian} that also fits our framework. The new formulation enables us to propose a fast and scalable primal dual algorithm for our model, but also for the one of \cite{dong2015laplacian} that was missing from the literature. Our experiments suggest that when sparse graphs are to be learned, but connectivity is crucial, our model is expected to outperform the current state of the art. 

We hope not only that our solution will be used for many applications that require good quality graphs, but also that our framework will trigger defining new graph learning models targeting specific applications.



\subsubsection*{Acknowledgements}
The author would like to especially thank Pierre Vandergheynst and Nikolaos Arvanitopoulos for their constructive comments on the organization of the paper and the experimental evaluation. He is also grateful to the authors of \textcite{dong2015laplacian} for sharing their code, to Nathanael Perraudin and Nauman Shahid for discussions when developing the initial idea, and to Andreas Loukas for his comments on the final version. 

\printbibliography

\newpage
\appendix


\begin{table*}
\caption{Different Types of Smooth Signals.}
\label{tab:smooth_filters}
\ra{1.8}
\centering
\begin{tabular}{@{}lll@{}}
\toprule
Concept & Model & Graph filter\\
\midrule
Tikhonov & $X = \arg\min_X \frac{1}{2}\|X-X_0\|_F^2 + \frac{1}{\alpha}\tr{X^\top LX}$ & $ g(\lambda) = \frac{1}{1+\alpha\lambda}$ \\
Generative model & $X\sim{\cal N}\left(0, L^\dagger\right)$ &$ g(\lambda) = \begin{cases} {\frac{1}{\sqrt{\lambda}}} &\mbox{if } \lambda > 0\\
0 &\mbox{if } \lambda=0 \end{cases}$\\
Heat diffusion & $X = \exp\left(-\alpha L\right)X_0$ & $ g(\lambda) = \exp(-\alpha\lambda)$\\
\bottomrule
\end{tabular}
\end{table*}

\begin{figure}
\centering
\includegraphics[scale = .33]{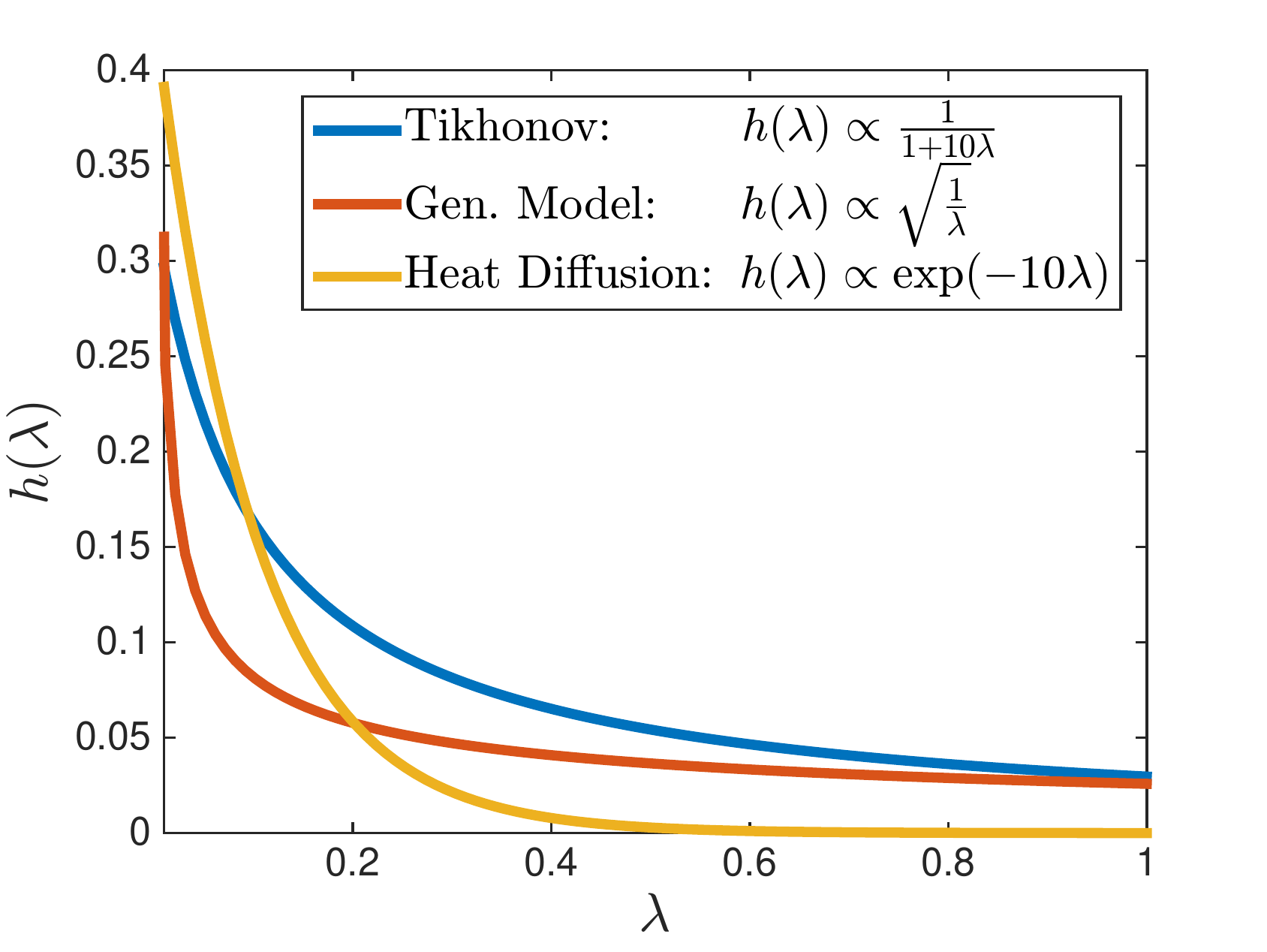}
\caption{The filters of Table \ref{tab:smooth_filters} for $\alpha = 10$.}
\end{figure}

\begin{figure}
\centering
\includegraphics[scale=.7, clip=true, trim= 80 36 72 24]{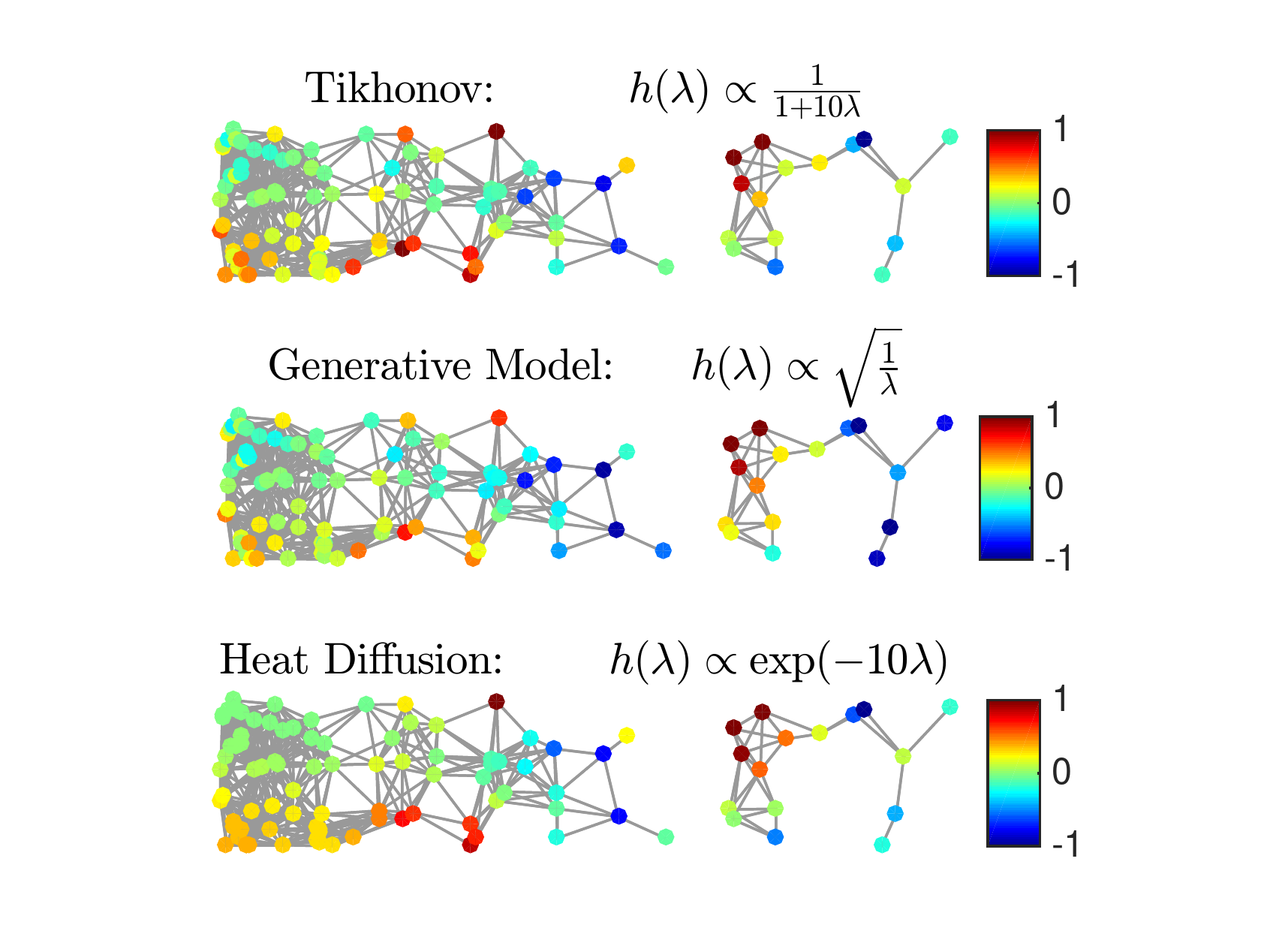}
\caption{Different smooth signals on the Non Uniform graph used for our artificial data experiments. All signals are obtained by smoothing the same initial $x_0\sim {\cal N}(0,I)$ with three different filters. This instance of the graph is disconnected with 2 components.}
\label{fig:non_uniform_signals}
\end{figure}

\section{Derivations and proofs}
\subsection{Detailed explanation of eq. \eqref{eq:smoothness_sparsity}}
\begin{align*}
\|W\circ Z\|_1&=\sum_{i=1}^m\sum_{j=1}^m W_{ij}\|x_i-x_j\|_2^2 \\
&=\sum_{i=1}^m\sum_{j=1}^m (x_i - x_j)^\top W_{ij}(x_i-x_j)  \\
&=2\sum_{i=1}^m\sum_{j=1}^m x_i^\top W_{ij}x_i -2\sum_{i=1}^m\sum_{j=1}^m x_i^\top W_{ij}x_j \\
&=2\sum_{i=1}^m x_i^\top x_i\sum_{j=1}^m W_{ij} -2\tr{X^\top W X}\\
&=2\tr{X^\top D X} -2\tr{X^\top W X} \\
&= 2\tr{X^\top LX},
\end{align*}
where $D$ is the diagonal matrix with elements $D_{ii} = \sum_i W_{ij}$.

\subsection{Proof of proposition \ref{proposition2}}
\begin{proof}
We change variable $\tilde W = W/\gamma$ to obtain 
\begin{align*}  &F(Z,\alpha, \beta) = \\
&~= \gamma\argmin_{\tilde W} \|\gamma \tilde W\circ Z\|_{1,1} - \alpha \ones^\top \log(\gamma \tilde W\ones) + \beta \|\gamma \tilde W\|_F^2\\
 &~= \gamma\argmin_{\tilde W} \gamma\| \tilde W\circ Z\|_{1,1} - \alpha \ones^\top \log( \tilde W\ones) + \beta \gamma^2\| \tilde W\|_F^2\\
 &~= \gamma\argmin_{\tilde W} \| \tilde W\circ Z\|_{1,1} - \frac{\alpha}{\gamma} \ones^\top \log( \tilde W\ones) + \beta \gamma\| \tilde W\|_F^2\\
 &~=\gamma F\left(Z, \frac{\alpha}{\gamma}, \beta\gamma\right),
\end{align*}
where we used the fact that $\log(\gamma \tilde W\ones) = \log(\tilde W\ones)+\constant(W)$. The second equality is obtained from the first one for $\gamma=\alpha$.
\end{proof}

\subsection{Proof of proposition \ref{proposition3}}
\begin{proof}
For equation \eqref{eq:xiaowen_prop1}
\begin{align*}
&H(Z + \gamma, \alpha, s) = \\
&=\argmin_{W\in{\cal W}_m} \|W\circ Z + \gamma W\|_{1,1} + \alpha \|W\|_F^2 + \alpha \|W\ones\|^2\\
& ~~~~~~\st, \qquad \|W\|_{1,1}=s\\
&=\argmin_{W\in{\cal W}_m} \|W\circ Z\|_{1,1} + \gamma \|W\|_{1,1} + \alpha \|W\|_F^2 + \alpha \|W\ones\|^2\\
& ~~~~~~\st, \qquad \|W\|_{1,1}=s\\
&=\argmin_{W\in{\cal W}_m} \|W\circ Z\|_{1,1} + \gamma s + \alpha \|W\|_F^2 + \alpha \|W\ones\|^2\\
& ~~~~~~\st, \qquad \|W\|_{1,1}=s\\
&=H(Z, \alpha, s),
\end{align*}
because $\|a+b\|_1 = \|a\|_1+\|b\|_1$ for positive $a$, $b$. 

For equation \eqref{eq:xiaowen_prop2} we change variable in the optimization and use $\tilde W = W/\gamma$ to obtain
\begin{align*}
&H\left(Z, \alpha, s\right) = \\
&~=\gamma \argmin_{\tilde W\in{\cal W}_m} \|\gamma\tilde W\circ Z\|_{1,1} + \alpha \|\gamma\tilde W\|_F^2 + \alpha \|\gamma\tilde W\ones\|^2\nonumber\\
& ~~~~~~\st, \qquad \|\gamma\tilde W\|_{1,1}=s\\
&~=\gamma \argmin_{\tilde W\in{\cal W}_m} \gamma\|\tilde W\circ Z\|_{1,1} + \gamma^2\alpha \|\tilde W\|_F^2 + \gamma^2\alpha \|\tilde W\ones\|^2\nonumber\\
& ~~~~~~\st, \qquad \|\tilde W\|_{1,1}=\frac{s}{\gamma}\\
&~=\gamma \argmin_{\tilde W\in{\cal W}_m} \|\tilde W\circ Z\|_{1,1} + \gamma\alpha \|\tilde W\|_F^2 + \gamma\alpha \|\tilde W\ones\|^2\nonumber\\
& ~~~~~~\st, \qquad \|\tilde W\|_{1,1}=\frac{s}{\gamma}\\
&~=\gamma H\left(Z, \alpha\gamma, \frac{s}{\gamma}\right).
\end{align*}
The second equality follows trivially for $\gamma  = s$.

\end{proof}



\section{Optimization details and algorithm for model of \cite{dong2015laplacian}}

To obtain Algorithm 1 (for our model), we need the following:
\begin{align*}
&K=S ~~~(\|S\|_2 = \sqrt{2(m-1)})\\
&\prox_{\lambda f_1}(y) = \max(0, y-\lambda z),\\
&\prox_{\lambda f_2}(y) = \frac{y_i+\sqrt{y_i^2+4\alpha\lambda}}{2},\\
&\nabla f_3(w) = 2\beta w,\\
&\zeta = 2\beta \qquad\text{(Lipschitz constant of gradient of $f_3$)},
\end{align*}
where $m$ is the number of nodes of the graph.

To obtain Algorithm 2 (for model by \cite{dong2015laplacian}), we need the following:
\begin{align*}
&K=2\ones ~~~(\|2\ones\|_2 = 2\sqrt{m(m-1)/2})\\
&\prox_{\lambda f_1}(y) = \max(0, y-\lambda z),\\
&\prox_{\lambda f_2}(y) = s,\\
&\nabla f_3(w) = \alpha(4w+2S^\top Sw),\\
&\zeta = 2\alpha(m+1) \text{ (Lipschitz constant of gradient of $f_3$)}.
\end{align*}

\begin{algorithm}
\caption{Primal dual algorithm for model \cite{dong2015laplacian}.}
\label{alg:xiaowens}
\begin{algorithmic}[1]
\State \textbf{Input:} $z, \alpha, s, w^0\in{\cal W}_v$, $c^0\in\mathbb{R}_+$, $\gamma$, tolerance $\tol$
\For{$i = 1,\dots,i_{max}$}
\State $ y^i = w^i - \gamma (2\alpha(2w^i + S^\top Sw^i) + 2c^i)$
\State $\bar y^i = c^i + \gamma (2\sum_jw^i_j)$
\State $p^i = \max(0, y^i-2\gamma z)$
\State $\bar p^i = \bar y^i -  \gamma s$
\State $q^i = p^i - \gamma (2\alpha(2p^i + S^\top Sp^i) + 2p^i)$
\State $\bar q^i = \bar p^i + \gamma (2\sum_jp^i_j)$
\State $w^i = w^i - y^i + q^i$;
\State $c^i = c^i - \bar y^i + \bar q^i$;
\If{$\|w^i-w^{i-1}\|/\|w^{i-1}\|<\tol$ \textbf{and} \\
\qquad\quad $|c^i-c^{i-1}|/|c^{i-1}|<\tol$}
\State \textbf{break}
\EndIf
\EndFor
\end{algorithmic}
\end{algorithm}


\begin{table*}[t]
\centering
\footnotesize
\caption{Performance of Different Algorithms on Artificial Data. Each setting has a random graph with 100 nodes and \vassilis{\textbf{100}} smooth signals from 3 different smoothness models and added $10\%$ noise. Results averaged over 20 random graphs for each setting. F-measure: the bigger the better (weights ignored). Edge and degree distances: the lower the better. For relative $\ell-1$ distances we normalize s.t. $\|w\|_1=\|w_0\|_1$. For relative $\ell-2$ distances we normalize s.t. $\|w\|_2=\|w_0\|_2$. Baseline: for F-measure, the best result by thresholding $\exp(-d^2)$. For edge and degree distances we use $\exp(-d^2/2\sigma^2)$ without thresholding.} \label{tab:artificial_100}
\begin{tabular}{@{}rcccccccccccc@{}}\toprule
& \phantom{}& \multicolumn{3}{c}{Tikhonov} & \phantom{}& \multicolumn{3}{c}{Generative Model} & \phantom{} & \multicolumn{3}{c}{Heat Diffusion}\\ \cmidrule{3-5} \cmidrule{7-9} \cmidrule{11-13}
&& \textbf{base} & \textbf{\hspace{-5pt}Dong etal\hspace{-5pt}} & \textbf{Ours} &&  \textbf{base} & \textbf{\hspace{-5pt}Dong etal\hspace{-5pt}} & \textbf{Ours} &&  \textbf{base} & \textbf{\hspace{-5pt}Dong etal\hspace{-5pt}} & \textbf{Ours} \\
\midrule
\textbf{Rand. Geometric}\\\cline{1-1}
F-measure &&           
0.667  & 0.860    & \textbf{0.886}  &&
0.671 & 0.836 & \textbf{0.858} &&
0.752 & 0.837 & \textbf{0.848} \\  
edge $\ell$-1 
&&      0.896  & 0.414    & \textbf{0.364} && 0.851 & 0.487 & \textbf{0.468} && 0.620 & 0.526 & \textbf{0.451}   \\
edge $\ell$-2 
&&      0.700  & 0.430    & \textbf{0.390}  && 0.692 & 0.494 & \textbf{0.477} && 0.582 & 0.535 & \textbf{0.471} \\
degree $\ell$-1
&&    0.158  & 0.151    & \textbf{0.080} && 0.268 & 0.159 & \textbf{0.128} && 0.216 & 0.225 & \textbf{0.143}  \\
degree $\ell$-2 
&&    0.707  & 0.179    & \textbf{0.095} && 0.679 & 0.193 & \textbf{0.145} && 0.479 & 0.264 & \textbf{0.177} \\
\textbf{Non Uniform}\\\cline{1-1}
F-measure &&     0.674  & \textbf{0.821}    & 0.817  && 0.650 & \textbf{0.779} & 0.774 && 0.763 & \textbf{0.835} & 0.827 \\
edge $\ell$-1 && 0.847  & 0.547    & \textbf{0.480} && 0.931 & 0.711 & \textbf{0.673} && 0.612 & 0.583 & \textbf{0.491} \\
edge $\ell$-2 &&     0.724  & 0.545    & \textbf{0.462} && 0.784 & 0.673 & \textbf{0.624} && 0.565 & 0.598 & \textbf{0.464}\\
degree $\ell$-1 &&   0.167  & 0.190    & \textbf{0.075} && 0.241 & 0.204 & \textbf{0.139} && 0.235 & 0.257 & \textbf{0.132}\\
degree $\ell$-2 &&   0.605  & 0.228    & \textbf{0.099} && 0.614 & 0.261 & \textbf{0.187} && 0.433 & 0.325 & \textbf{0.164}\\
\textbf{\ER}\\\cline{1-1}
F-measure &&      0.293  & 0.595    & \textbf{0.676}&& 0.207 & 0.473 & \textbf{0.512}
&& 0.358 & 0.595 & \textbf{0.619} \\
edge $\ell$-1 &&  1.513  & 0.837    & \textbf{0.798} && 1.623 & 1.113 & \textbf{1.090} && 1.401 & \textbf{0.896} & 0.899 \\
edge $\ell$-2 &&     1.086  & 0.712    & \textbf{0.697} && 1.129 & 0.896 & \textbf{0.888}
&& 1.045 & 0.767 & \textbf{0.759} \\
degree $\ell$-1 &&   0.114  & 0.129    & \textbf{0.084}&& 0.135 & 0.146 & \textbf{0.114}&& 0.185 & \textbf{0.182} & 0.184 \\ 
degree $\ell$-2 &&   0.932  & 0.202    & \textbf{0.116} && 1.053 & 0.227 & \textbf{0.185} && 0.875 & \textbf{0.241} & 0.276 \\
\textbf{Barab\'asi-Albert}\\
\cline{1-1}
F-measure &&        0.325  & 0.564    & \textbf{0.636}&& 0.357 & 0.588 & \textbf{0.632}&& 0.349 & 0.631 & \textbf{0.711} \\
edge $\ell$-1 &&    1.541  & 0.939    & \textbf{0.885}&& 1.513 & 0.940 & \textbf{0.914}&& 1.473 & 0.843 & \textbf{0.774}  \\
edge $\ell$-2 &&    1.073  & 0.802    & \textbf{0.761}&& 1.052 & 0.808 & \textbf{0.773}&& 1.049 & 0.732 & \textbf{0.672}  \\
degree $\ell$-1& &  0.225  & 0.309    & \textbf{0.145}&& 0.243 & 0.311 & \textbf{0.229} && 0.281 & 0.336 & \textbf{0.181} \\ 
degree $\ell$-2 &&  0.560  & 0.378    & \textbf{0.281} && 0.563 & 0.386 & \textbf{0.350}&& 0.570 & 0.429 & \textbf{0.319} \\
\bottomrule
\end{tabular}
\end{table*}

\section{More real data experiments}
\begin{figure}
\centering
\includegraphics[scale=.4]{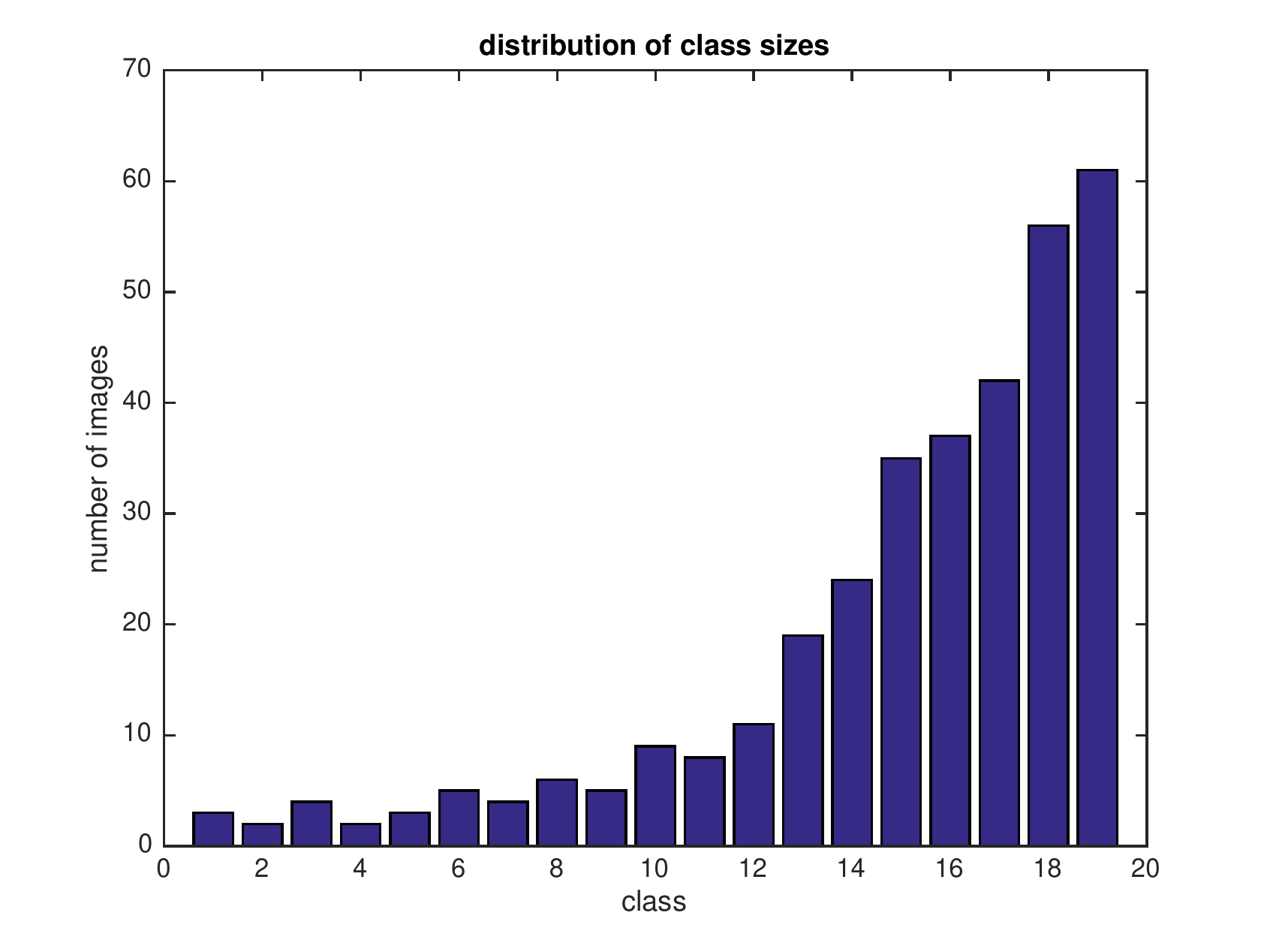}
\caption{Distribution of class sizes for one of the random instances of the COIL 20 experiments.}
\label{fig:coil_sizes}
\end{figure}

\subsection{Learning the graph of COIL 20 images}
We randomly sample the classes so that the average size increases non-linearly from around $3$ to around $60$ samples per class. The distribution for one of the instances of this experiment is plotted in fig. \ref{fig:coil_sizes}. We sample from the same distribution {20 times} and measure the average performance of the models for different graph densities. For each of the graphs, we run standard spectral clustering (as in the work of \cite{ng2002spectral} but without normalizing the Laplacian) with k-means {100 times}. For label propagation we choose {100 times} a different subset of $50\%$ known labels. We set a baseline by using the same techniques with a k-Nearest neighbors graph (k-NN) with different choices of $k$.

\begin{figure*}
    \centering
    \begin{subfigure}[b]{0.32\textwidth}
        \includegraphics[scale=.32]{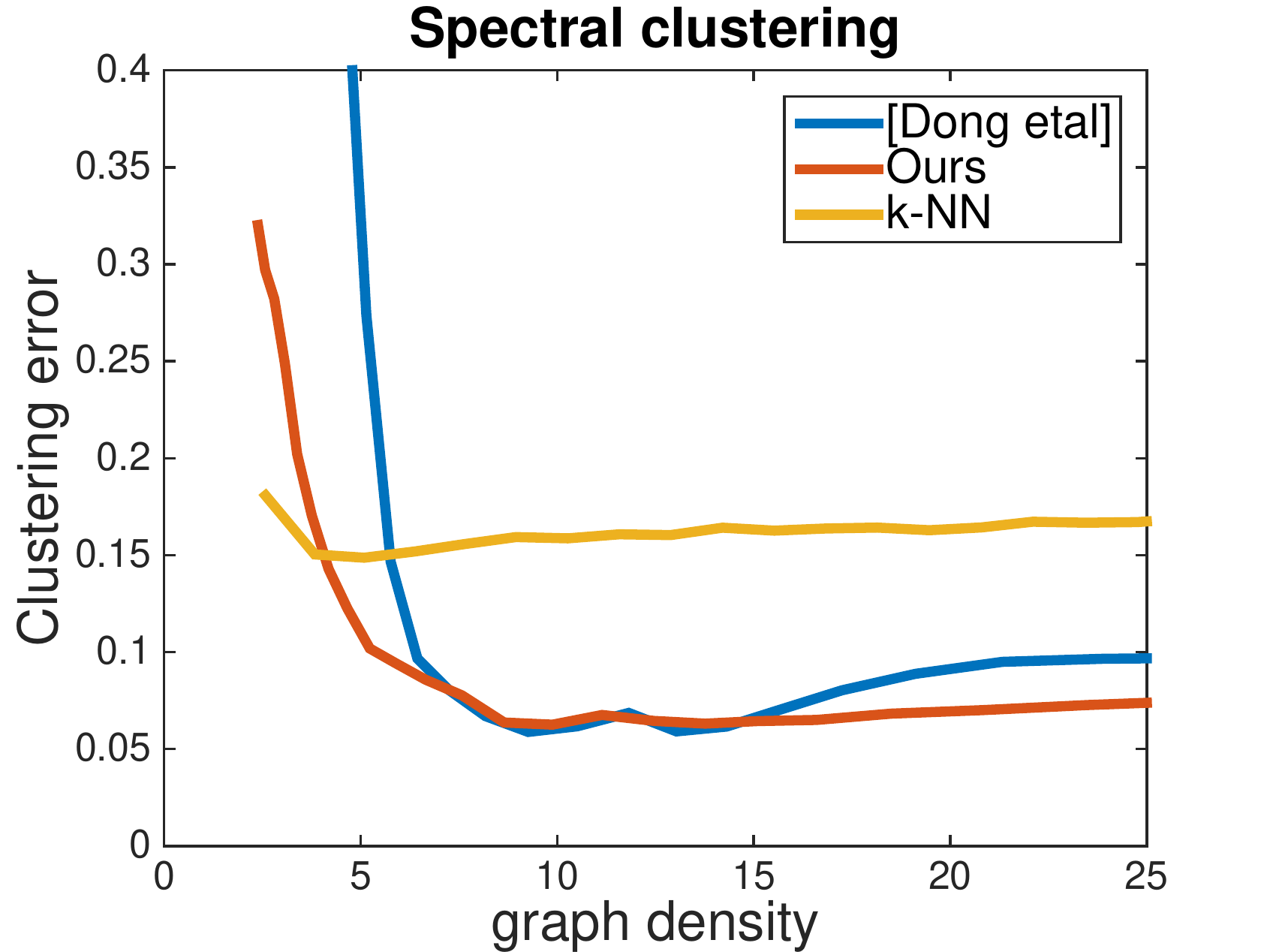}
    \end{subfigure}
    \begin{subfigure}[b]{0.32\textwidth}
        \includegraphics[scale=.32]{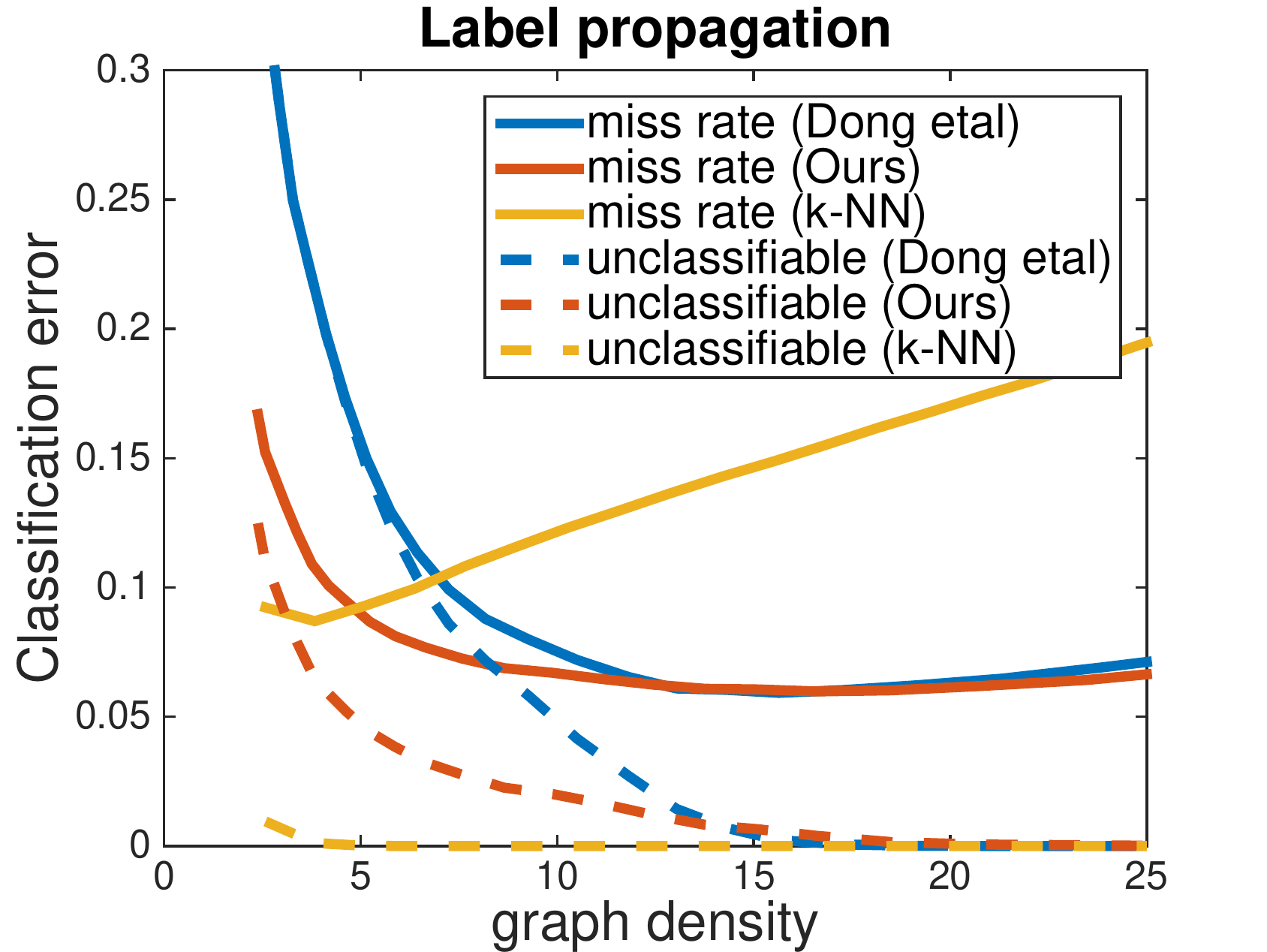}
    \end{subfigure}
    \begin{subfigure}[b]{0.32\textwidth}
        \includegraphics[scale=.32]{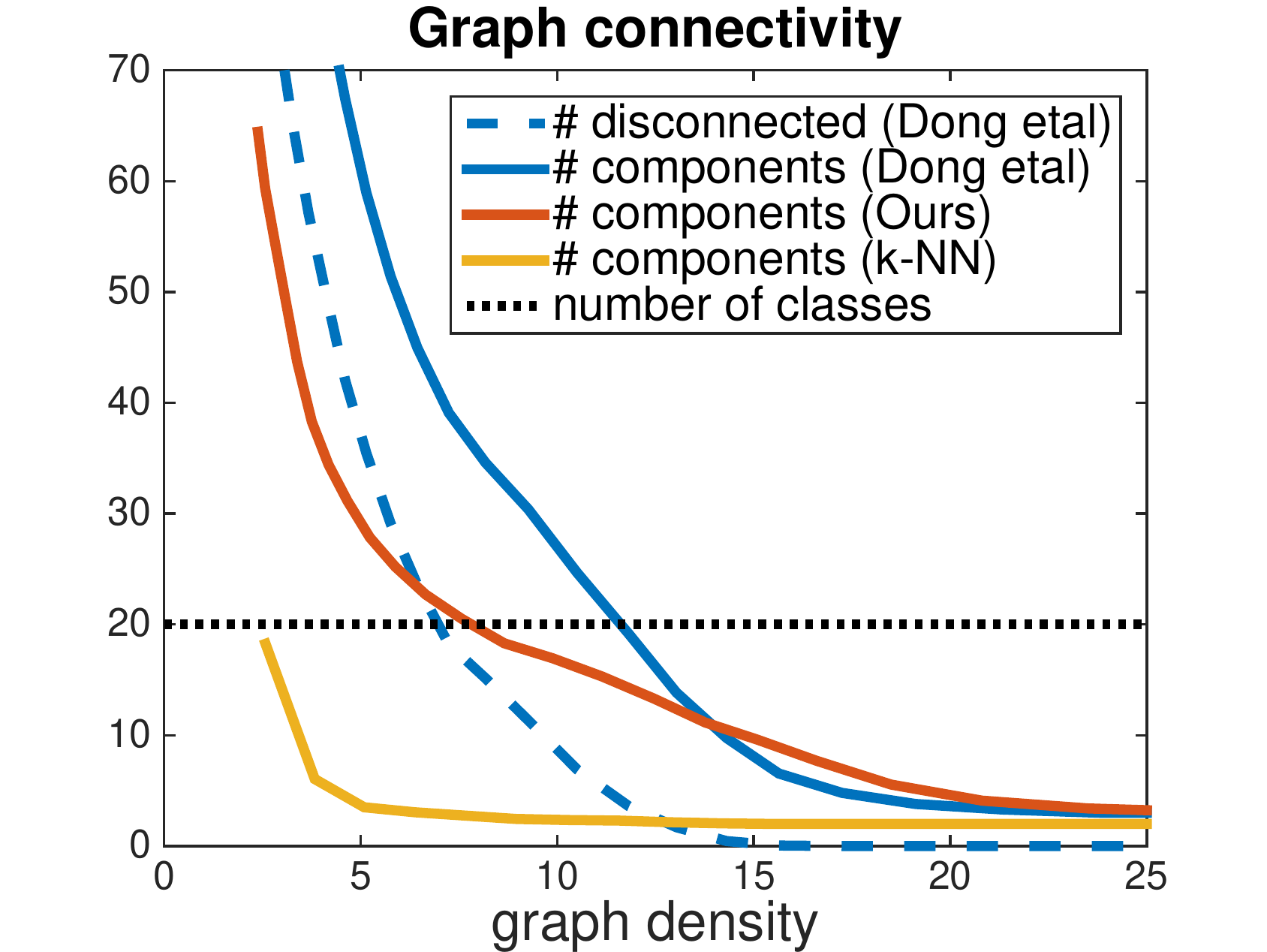}
    \end{subfigure}
    \caption{Graph learned from non-uniformly sampled images from COIL 20. Average over 20 different samples from the same non-uniform distribution of images. \textbf{Left}: Clustering quality. \textbf{Middle}: Label propagation quality. Dashed lines are the number of nodes  in components without labeled nodes. \textbf{Right}: Number of disconnected components and number of disconnected nodes (Our model and k-NN have no disconnected nodes).}\label{fig:COIL_20}
\end{figure*}

In Fig. \ref{fig:COIL_20} we plot the behavior of different models for different density levels. The horizontal axis is the average number of non-zero edges per node.

The dashed lines of the middle plot denote the number of nodes contained in components without labeled nodes, that can not be classified.

\end{document}